\documentclass[runningheads]{llncs}

\usepackage{eccv}

\usepackage{eccvabbrv}

\usepackage{graphicx}
\usepackage{booktabs}
\usepackage{verbatim}
\usepackage{placeins}
\usepackage{tcolorbox}
\tcbuselibrary{breakable}
\usepackage[accsupp]{axessibility}  

\usepackage{hyperref}

\usepackage{orcidlink}

\usepackage{wrapfig} 
\usepackage{xcolor}
\usepackage[table]{xcolor}
\definecolor{second}{RGB}{200,230,255}     
\definecolor{best}{RGB}{127,155,244}   

\renewcommand{\paragraph}[1]{\vspace{0.5em}\noindent\textbf{#1}}

\begin{document}

\title{Think While You Map:\\Asynchronous Vision-Language Agents for Incremental 3D Scene Graphs} 

\titlerunning{Think While You Map}

\author{Deniz Bickici\inst{1,3}\orcidlink{0009-0001-4827-6933},
Michael Pabst\inst{1,3}\orcidlink{0000-0002-7985-4652},
Shohei Mori\inst{1}\orcidlink{0000-0003-0540-7312}, Dieter Schmalstieg\inst{1,2}\orcidlink{0000-0003-2813-2235}}

\authorrunning{D.~Bickici et al.}

\institute{University of Stuttgart, Germany \and Graz University of Technology, Austria \and IMPRS-IS, Germany\\
  \email{deniz.bickici@visus.uni-stuttgart.de}}

\maketitle

\begin{abstract}
  Open-vocabulary 3D scene graph methods typically operate in two stages: first reconstruct, then enrich with vision-language models, leaving the graph unqueryable during exploration. We argue that this sequential coupling is unnecessary and propose an asynchronous architecture in which lightweight online mapping runs concurrently with heavyweight semantic refinement. A probabilistic voxel-based backbone maintains stable object identities incrementally, while background VLM agents progressively enrich the graph. This framework resolves duplicate object tracks through semantic loop closure, attaches fine-grained visual attributes and derives spatial relations between objects. A multi-target frame scheduler amortizes VLM cost by selecting a small set of informative frames that jointly cover multiple targets. The resulting scene graph is queryable during exploration and grows in semantic richness over time. Our method matches or outperforms existing open-vocabulary 3D scene graph methods on semantic segmentation (ScanNet, Replica) and surpasses the prior state-of-the-art across three visual grounding benchmarks (Sr3D+, Nr3D, ScanRefer) by 15.3 to 18.8 A@0.25.
  Project page: \url{https://denizbickici.github.io/thinkgraphs/}
\end{abstract}

\begin{figure}[tb]
    \centering
    \includegraphics[width=\linewidth]{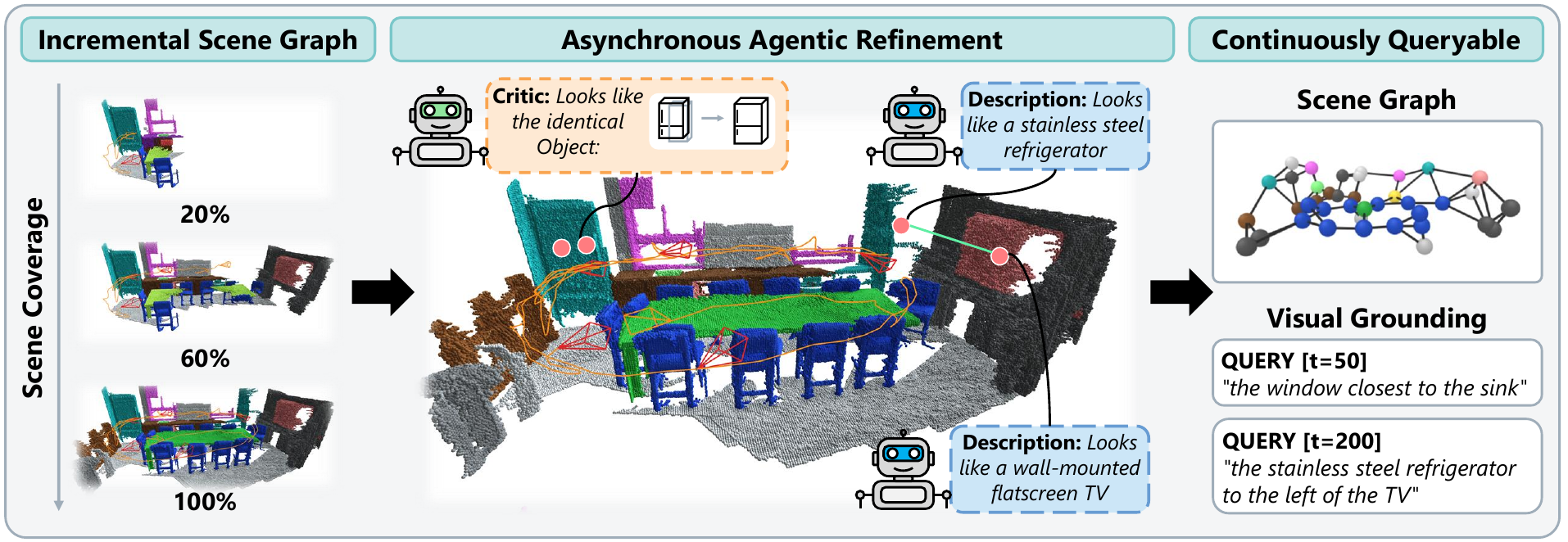}
    \caption{\textbf{ThinkGraphs} decouples lightweight online mapping from heavyweight VLM reasoning, keeping the scene graph queryable during exploration. Asynchronous background agents refine the graph without blocking the mapping loop: a Critic Agent detects and merges fragmented object tracks, and a Description Agent injects fine-grained visual attributes, enabling complex grounding queries (e.g., ``the stainless steel refrigerator left of the TV'') throughout exploration.}
    \label{fig:teaser}
\end{figure}

\section{Introduction}
\label{sec:intro}

Understanding the semantics of 3D environments is fundamental for embodied AI, robotics, and augmented reality.
An agent must recognize objects, maintain their identities over time, infer relationships and support language-based queries for grounding and planning, e.g., ``pick up the mug next to the laptop''. Scene graphs are a natural representation for this goal.
They compactly encode entities as nodes and their interactions as edges, enabling downstream reasoning about ``what is where'' and ``what is interacting with what.''

Recent work on 3D scene graphs has largely evolved along two complementary but currently disconnected directions:
Semantic-rich offline reconstruction prevents querying unless the whole process is complete, while real-time reconstruction supports querying but only with limited semantic information.
The former utilizes offline open-vocabulary systems and prioritizes rich labels, attributes, free-form descriptions, and functional relations~\cite{gu2024conceptgraphs, werby2024hierarchical, linok_bare_2025, zhang2025openvocabularya, mei_vocabularyfree_2025}, typically leveraging CLIP~\cite{radford2021learning} embeddings and vision-language reasoning.
These methods require completing instance-level reconstruction before enriching the scene with captions, labels, and inter-object relations.
The latter are real-time mapping systems that incrementally update object graphs but rely on closed sets of semantics or shallow feature matching, trading the semantic richness of recent large language models (LLM) or vision-language models (VLM) for responsiveness~\cite{maggio_clio_2024,hughes2022hydraa,wu2023incrementalb}.

The core technical challenge of combining these two paradigms lies in the instability of incremental tracking. Since objects are often fragmented or merged incorrectly under partial views and occlusions, it is impractical to query a VLM during exploration. Not only would this degrade the system's responsiveness, it would also waste massive computational resources on analyzing duplicate or unstable object tracks. Consequently, previous systems have to wait for a fully consolidated offline reconstruction before applying VLM reasoning.

Many semantically rich pipelines already build their geometric map incrementally~\cite{gu2024conceptgraphs, linok_bare_2025}. What they lack is this combination of a robust enough 3D scene representation to make on-the-fly enrichment worthwhile, and an efficient scheduling mechanism to make it affordable.

Our key observation is that semantic reasoning does not need to wait for a stable graph; it can actively contribute to stabilizing it. This behavior is enabled by two factors. First, rather than the pairwise greedy association used by prior methods, we accumulate per-voxel votes for each object track, producing substantially more robust object tracks, reducing the instability that motivated postponing semantics in the first place. Second, rather than brute-forcing VLM calls on every object at every frame, we treat VLM reasoning as a selective, asynchronous background process, one that not only enriches the graph with fine-grained attributes, but also actively repairs it by detecting and merging duplicate object tracks through semantic loop closure. 

We therefore propose an asynchronous architecture in which a lightweight online mapper runs concurrently with two background VLM agents: a Critic Agent that detects and merges duplicate object tracks through semantic loop closure, and a Description Agent that attaches fine-grained visual attributes via a multi-target frame scheduler that amortizes VLM cost across informative frames. The resulting scene graph grows in semantic richness as exploration continues, yet never blocks online operation. Our method thus occupies the middle ground between offline open-vocabulary reconstruction and real-time closed-set mapping. It builds the graph incrementally during exploration, enriching semantics asynchronously rather than at frame rate.

To demonstrate the effectiveness of our approach, we extensively evaluate it on open-vocabulary semantic segmentation and three visual grounding benchmarks.
In summary, we make the following contributions:
\begin{itemize}
    \item We introduce \textit{ThinkGraphs}, to our knowledge the first \textit{asynchronous} architecture for open-vocabulary 3D scene graph construction. It decouples lightweight online mapping from heavyweight VLM reasoning, meaning the graph remains queryable during exploration and is continually enriched without blocking it.
    
    \item We propose a VLM-in-the-loop mechanism for semantic loop closure that detects and merges duplicate object tracks caused by incremental association drift, analogous to geometric loop closure in SLAM.
    
    \item We introduce a multi-target frame scheduler that selects a small number of informative frames for the Description Agent, each depicting multiple undescribed objects, so that a single VLM call enriches several nodes at once.
    
    \item We demonstrate that our incremental method outperforms all batch-based open-vocabulary scene graph methods on semantic segmentation and surpasses the prior state-of-the-art across three visual grounding benchmarks (Sr3D+, Nr3D, ScanRefer) by 15.3 to 18.8 A@0.25, showing that deferring all semantics to a post-hoc stage is unnecessary.

\end{itemize}

\section{Related Work}
\label{sec:related-work}


\paragraph{Incremental 3D scene graphs.}
The incremental construction of 3D scene graphs focuses on updating structured spatial representations as new observations arrive.
Kimera~\cite{rosinol_kimera_2020a, rosinol_kimera_2021a} first demonstrated real-time metric-semantic mapping with hierarchical spatial layers, a design later extended by Hydra~\cite{hughes2022hydraa}, which pioneered incremental scene graphs by integrating signed distance fields with hierarchical graphs of objects, places, and rooms.
S-Graphs+~\cite{bavle_sgraphs_2023} introduced a factor-graph formulation for hierarchical optimization, while SceneGraphFusion~\cite{wu_scenegraphfusion_2021}
fuses geometric over-segmentation with a Graph Neural Network (GNN) for incremental semantic prediction.
Wu et al.~\cite{wu2023incrementalb} propose a sparse RGB-only variant that combines SLAM-based point tracking with multi-view feature aggregation. Further incremental methods either reuse temporal history~\cite{feng2025history} or incorporate knowledge from past observations~\cite{renz2025integrating} to enhance scene understanding. All of these methods remain restricted to closed-set vocabularies, limiting their
applicability in open-world settings. More recently, Clio~\cite{maggio_clio_2024} introduced language interfaces into incremental mapping, averaging open-set features across temporal object tracks to form clusters relevant to the task, although it was limited to a predefined task list.

Compared to prior incremental scene graph mappers, our goal is to keep the graph open-vocabulary and queryable while exploration is ongoing. We achieve this goal by decoupling lightweight online tracking from heavyweight semantic enrichment: the mapper maintains stable object identities and geometry, while background VLM agents progressively refine labels and attributes without blocking the main loop; spatial relations are updated deterministically from 3D geometry.


\paragraph{Batch-processing of scene graphs.}
Many open-vocabulary 3D systems represent semantics as language-aligned features stored in dense maps or spatial memories~\cite{peng_openscene_2023,yamazaki_openfusion_2024,huang_visual_2023a,shafiullah_clipfields_2023,chen_openvocabulary_2023,liu_dynamem_2025a}.
In contrast, batch scene graph pipelines commit these signals to discrete object nodes and relations, enabling compositional and relational queries.

ConceptGraphs~\cite{gu2024conceptgraphs} pioneered open-vocabulary 3D scene graphs by combining CLIP~\cite{radford2021learning} embeddings with SAM2~\cite{ravi_sam_2024} masks to construct 3D objects.
Subsequent works extended this paradigm by adding hierarchical multi-level reasoning across rooms and floors~\cite{werby2024hierarchical}, object affordances~\cite{zhang2025openvocabularya}, and relational question answering~\cite{linok_bare_2025}. Open3DSG~\cite{koch2024open3dsg} takes a learning-based approach by distilling 2D vision-language features into a graph neural network, producing language-aligned node and edge embeddings for open-vocabulary object and pairwise relationship prediction. PoVo~\cite{mei_vocabularyfree_2025} takes a different approach, operating on geometric superpoints rather than per-frame masks and merging them into 3D instances through combined semantic and mask coherence. 

While some of these methods build their geometric map incrementally~\cite{gu2024conceptgraphs,linok_bare_2025}, all defer semantic enrichment to a post-hoc stage and therefore require a complete reconstruction before the graph becomes queryable. Our method instead exposes an intermediate graph throughout exploration and improves it incrementally: Semantic refinement is asynchronous and revisitable, so the system can answer queries early with coarse semantics and later update the same nodes or edges as better evidence emerges.


\paragraph{VLM reasoning in 3D scene understanding.   }
Large vision-language models bring contextual reasoning capabilities that go beyond the embedding similarity used for object association and labeling.
In open-vocabulary scene graphs, ConceptGraphs~\cite{gu2024conceptgraphs} queries a VLM to caption each object node and an LLM to infer spatial edges, while BBQ~\cite{linok_bare_2025} formulates object retrieval as deductive reasoning over a scene graph structure using an LLM.

For 3D visual grounding, LLM-Grounder~\cite{yang_llmgrounder_2024} decompose complex natural-language queries with an LLM into sub-goals and iteratively invokes a 3D visual grounder (e.g., OpenScene~\cite{peng_openscene_2023} or LERF~\cite{kerr_lerf_2023}) to localize targets in reconstructed scenes.
SQA3D~\cite{ma_sqa3d_2022} benchmarks situated question answering that requires joint spatial and common-sense reasoning on 3D scans.
Effective visual grounding in these pipelines often relies on visual prompting strategies such as Set-of-Mark~\cite{yang2023setofmark}, which overlays numeric tags on segmented image regions so that a VLM can refer to specific objects without fine-tuning.

Existing 3D mapping methods generally employ VLM/LLM reasoning in one of two ways: post-hoc enrichment after reconstruction, or synchronous query-time inference that blocks execution until the model responds. Our work targets the missing middle ground: We treat VLM reasoning as a background process that runs continuously, is scheduled selectively, and integrates results when available. This makes semantics available during mapping while controlling the computation via coverage-based frame selection rather than per-object per-frame querying.


\paragraph{Object-level association and merging.  }
A key challenge in incremental scene graph mapping is maintaining consistent object identities as observations accumulate over time.
Prior work addresses this through object-level SLAM~\cite{mccormac_fusion_2018}, multi-object tracking and fusion~\cite{runz2018maskfusion,runz2017cofusion}, and semantic object-based loop closure~\cite{lin2021topologyb,liu2019global}. However, association remains challenging in the presence of perceptual ambiguity and viewpoint changes.

In open-set mapping, ROMAN~\cite{peterson2025romana} aligns object submaps via graph-theoretic data association but assumes object nodes are already formed, while OpenVox~\cite{yinan_openvox_2025} formulates incremental instance association as probabilistic voxel inference and Octree-Graph~\cite{wang2025openvocabulary} proposes chronological group-wise merging with informative feature selection.

Most scene graph systems, whether incremental~\cite{hughes2022hydraa,wu_scenegraphfusion_2021} or batch-based~\cite{gu2024conceptgraphs,mei_vocabularyfree_2025}, rely on greedy pairwise association in learned embeddings (e.g., CLIP~\cite{radford2021learning} or DINO~\cite{oquab_dinov2_2024}) and 3D geometric overlap.
Such greedy strategies involve making early, local commitments based on limited evidence, which result in the accumulation of false negatives that lead to fragmented object tracks~\cite{hughes2022hydraa,linok_bare_2025}.
BBQ~\cite{linok_bare_2025} addresses this by consolidating duplicates in a periodic merge step, but repeated instances and viewpoint changes can still cause missed associations.

In contrast, our method couples probabilistic instance tracking with a VLM-based semantic loop closure mechanism that explicitly detects and corrects fragmented object tracks during online operation (\cref{subsec:async}).

\section{Method}
\label{sec:method}

\begin{figure}[tb]
  \centering
  \includegraphics[width=\linewidth]{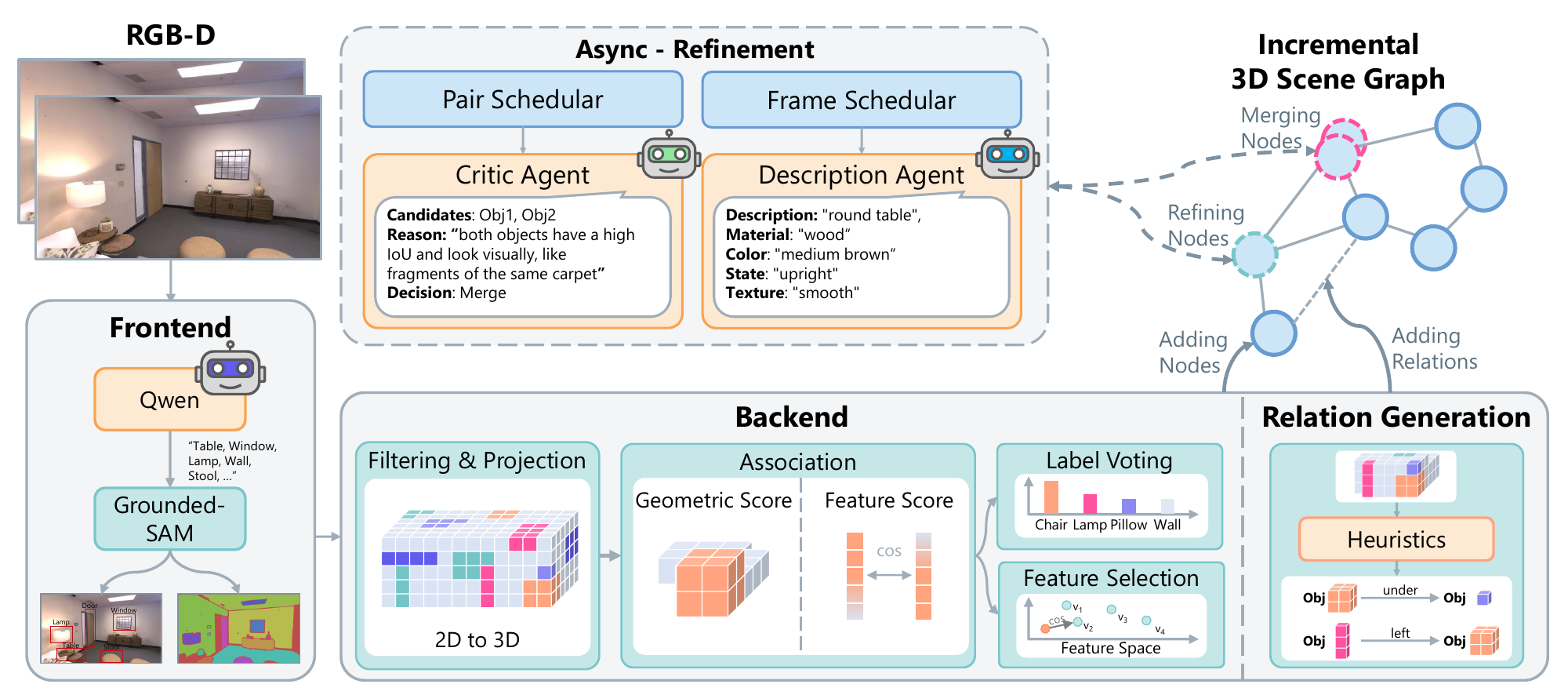}
  \caption{\textbf{Method Overview.} (i) The frontend extracts grounded instances from each RGB-D frame (\cref{subsec:frontend}). (ii) The backend associates them into persistent 3D tracks with probabilistic voxel scoring and derives spatial edges deterministically from 3D geometry (\cref{subsec:backend}). (iii) Two asynchronous VLM agents, a Critic Agent for semantic loop closure and a Description Agent for attribute enrichment, progressively refine the graph without blocking online operation (\cref{subsec:async}).}
  \label{fig:method}
\end{figure}

We introduce ThinkGraphs, an incremental open-vocabulary 3D scene graph pipeline that fuses geometric and visual evidence from sequentially posed RGB-D frames
$\{(I_t^{\text{rgb}},\, I_t^{\text{depth}},\, \theta_t)\}_{t=1}^{N}$ (\cref{fig:method}), where $\theta_t$ is the camera pose.
Our system incrementally constructs a scene graph $G_t = (\mathcal{V}_t, \mathcal{E}_t)$, where each node $v_i \in \mathcal{V}_t$ corresponds to a confirmed object track $T_i$, represented by a 3D point cloud $\mathbf{P}_i$,
a selected CLIP visual embedding $\mathbf{f}_i^\star$, a consensus label $\ell_i^\star$, and VLM-enriched attributes
$\mathcal{A}_i$. Edges $\mathcal{E}_t$ encode spatial predicates derived from 3D geometry.
Rather than requiring a complete reconstruction, the pipeline maintains a
continuously queryable graph throughout exploration via three components:
(i) a \textbf{frontend} that extracts grounded instance observations $\mathcal{O}_t$, (ii) a \textbf{backend} that lifts and associates them into persistent 3D object tracks, and (iii) \textbf{asynchronous VLM agents} that refine the scene graph without blocking the mapping loop.

\subsection{Frontend for grounded instance extraction}
\label{subsec:frontend}
Given an RGB image $I_t^{\text{rgb}}$ at time $t$, we query Qwen3-VL-2B-Instruct~\cite{bai_qwen3vl_2025}
to produce a set of noun prompts  $\mathcal{L}_t = \{ \ell_j^t \mid j=1,\dots,C_t \}$, as its context-aware prompts yield fewer missed detections than tag-based alternatives (\cref{tab:semseg_ablation}).
We feed these prompts into Grounded-SAM v2~\cite{ravi_sam_2024,ren2024grounded,liu_grounding_2025}, where
Grounding-DINO~\cite{liu_grounding_2025} predicts text-conditioned boxes for each prompt,
and SAM2~\cite{ravi_sam_2024} then refines each box into a 2D instance mask.
The frontend outputs a set of $K_t$ grounded instances $ \mathcal{O}_t = \{ (M_k^t,\, \ell_k^t,\, q_k^t) \}_{k=1}^{K_t}$, where each mask $M_k^t$ is paired with its matched prompt $\ell_k^t \in \mathcal{L}_t$ and confidence $q_k^t$. This ensures every mask carries a grounded label from the start, which our backend relies on for association and label assignment.

\subsection{Backend for 3D Association and Track Representation}
\label{subsec:backend}
We adopt OpenVox's~\cite{yinan_openvox_2025} probabilistic voxel-consistency 
formulation as our association backbone, as voxel-level voting accumulates evidence across multiple observations, filtering out noisy detections over time, whereas greedy pairwise matching commits to early associations that can accumulate drift.
The underlying voxel map is a sparse hash map that allocates cells only when RGB-D observations project points into them, so storage scales with observed surface rather than a prespecified volume.
We adapt it for our open-vocabulary setting by replacing its SBERT caption pipeline with CLIP text-label embeddings for track association, computed directly from the detector's label strings. These embeddings are invariant to viewpoint, unlike visual features. We therefore store visual embeddings separately for retrieval and grounding. Beyond the change of embedding, we introduce stricter observation filtering and confidence-weighted label voting to suppress noise and obtain more stable track representations.


\paragraph{Track scoring.}
Before association, we filter each 2D mask (erosion, subtraction of contained segments, area and confidence thresholds) and retain only the dominant 3D cluster via DBSCAN~\cite{ester_densitybased_1996}, which serves solely as a local per-mask denoising step.
We then maintain a sparse voxel map that continuously records the spatial occupancy history of each track. Given a new observation $O = (M_k^t,\, \ell_k^t,\, q_k^t) \in \mathcal{O}_t$, we compute an association score $S(T_i, O)$ for each candidate track $T_i$ using a weighted combination of geometric and feature similarities:
\begin{equation}
  S(T_i, O) = \lambda_{\text{geo}} \cdot S_{\text{geo}} + \lambda_{\text{feat}} \cdot S_{\text{feat}}.
  \label{eq:total_score}
\end{equation}
The geometric score $S_{\text{geo}}$ represents a probabilistic voxel-consistency vote in the intersecting volume:
\begin{equation}
  S_{\text{geo}}(T_i,O) = \frac{1}{|V(O)|}\sum_{v\in V(O)}\frac{c_{v,i}}{c_v},
  \label{eq:geo_score}
\end{equation}
where $c_{v,i}$ is the number of times track $T_i$ has been observed in voxel $v$, $c_v = \sum_j c_{v,j}$ is the total observation count across all tracks in that voxel, and $V(O)$ is the set of voxels occupied by observation $O$.
The feature score $S_{\text{feat}}$ is the cosine similarity between CLIP text embeddings of the incoming observation label and the track's current consensus label. With $\lambda_{\text{geo}}=0.8$ and $\lambda_{\text{feat}}=0.2$, the observation is merged into the highest-scoring track if $S \ge \tau_{\text{assoc}}$. Otherwise, a new tentative track is instantiated. To prevent noisy detections from cluttering the scene graph, these tentative object tracks are promoted to confirmed nodes only after accumulating at least $N_{\text{conf}}$ successful associations, effectively filtering out transient noise.


\paragraph{Confidence-weighted label voting.}
As observations accumulate, the same track may receive different label predictions across frames. To resolve these into a single consensus label, each track $T_i$ maintains a label histogram $\mathcal{H}_i$ over the labels observed so far. When a new observation with grounded label $\ell$ and detection confidence $q$ is associated to $T_i$, the histogram is updated as
\begin{equation}
    \mathcal{H}_{i}(\ell) \;\leftarrow\; \mathcal{H}_{i}(\ell) + q,
    \label{eq:label_vote}
\end{equation}
and the consensus label is the highest-scoring label:
\begin{equation}
    \ell_i^\star = \underset{\ell}{\arg\max}\; \mathcal{H}_{i}(\ell).
  \label{eq:aggregated_label}
\end{equation}%
Weighting votes by detection confidence, rather than uniform counting, ensures that high-quality detections naturally suppress noisy or ambiguous predictions.


\paragraph{Visual embedding selection.}
To support downstream retrieval and grounding tasks, each track $T_i$ aggregates a bank of CLIP visual embeddings $\mathcal{B}_i = \{\mathbf{f}_n^{\text{vis}}\}$, extracted from tight crops of its associated 2D segments. To ensure embedding quality, we apply two complementary filters.
First, distant or heavily occluded viewpoints produce low-quality embeddings and should be discarded. An \emph{adaptive area gate} maintains a running maximum segment area $A_i^\star$ over the track history and only admits semantic/CLIP updates whose visible 2D segment area $A_n$ satisfies $A_n \geq \rho_\text{area} \cdot A_i^\star$.
Second, we select a single representative embedding via \emph{text-guided scoring}. Relying solely on the largest crop is insufficient: For instance, a large view of a wall may still be dominated by occluding furniture, confounding the CLIP feature. Instead, we encode the track's consensus label $\ell_i^\star$ (\cref{eq:aggregated_label}) with the CLIP text encoder to obtain a semantic anchor $\mathbf{f}_{\ell_i^\star}^{\text{text}}$ and retrieve the most semantically aligned view:
\begin{equation}
    \mathbf{f}_i^{\star} = \mathbf{f}_{s^*}^{\text{vis}},
    \text{~where~}
    s^* = \arg\max_n\, {\mathbf{f}_n^{\text{vis}}}^\top \mathbf{f}_{\ell_i^\star}^{\text{text}}.
  \label{eq:text_guided}
\end{equation}
\paragraph{Spatial edges.} To support relational grounding queries, we derive directed spatial edges deterministically from the axis-aligned bounding boxes of confirmed object tracks, following a similar approach to BBQ~\cite{linok_bare_2025}, as spatial relations can be reliably determined from geometric heuristics. Each edge encodes a directional predicate (e.g., left, under) and pairwise distance between target and anchor. Horizontal directions are resolved relative to a virtual room-center anchor to ensure viewpoint invariance.

\begin{figure}[tb]
    \centering
    \includegraphics[width=\linewidth]{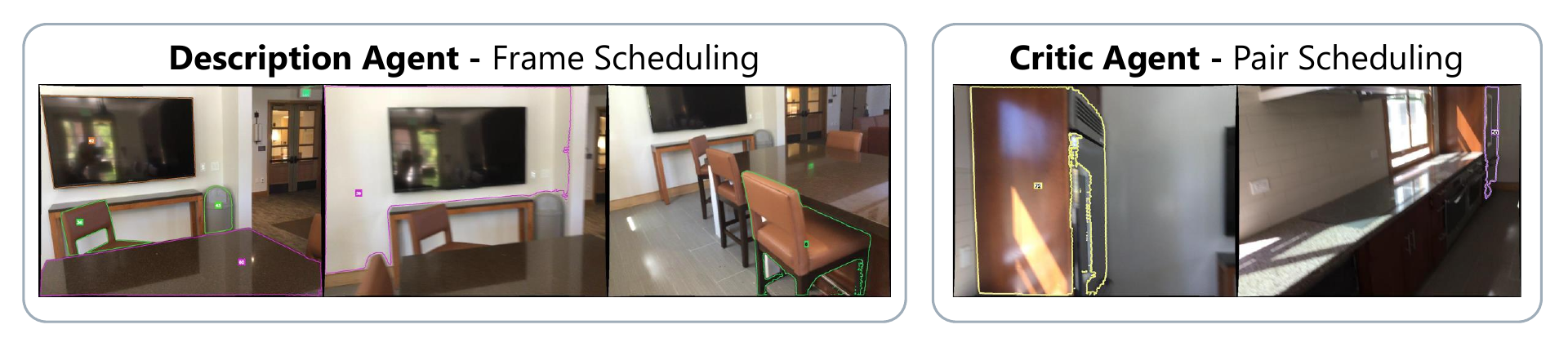}
    \caption{\textbf{VLM Scheduling.} Multi-target frame scheduling for the Description Agent (left) and pair scheduling for the Critic Agent (right), both using Set-of-Mark overlays.}
    \label{fig:schedular}
\end{figure}

\subsection{Asynchronous Agentic Refinement}
\label{subsec:async}
Enriching every node at every frame with a VLM would be prohibitively expensive, while deferring all enrichment to a post-hoc stage would mean that the system would not be queryable during exploration.
We therefore introduce an asynchronous refinement layer comprising two VLM background agents that refine the graph without blocking the online loop (\cref{fig:schedular}).
The Critic Agent detects and merges duplicate object tracks caused by incremental association errors, while the Description Agent attaches fine-grained visual attributes to each node.


\paragraph{Critic agent. } 
Similarly to pose drift in SLAM, incremental object associations accumulate errors over time, spawning duplicate or fragmented tracks. Prior consolidation methods~\cite{linok_bare_2025} rely on feature similarity and 3D overlap, but embeddings struggle to distinguish distinct instances of the same class or to re-identify single objects across varying viewpoints. Resolving such an ambiguity requires both fine-grained visual evidence and a broader scene context, motivating the use of a VLM as a verifier. We therefore introduce the Critic Agent: an asynchronous in-the-loop verifier that performs merge-only identity corrections in two stages. First, a lightweight \emph{candidate scheduling} step proposes high-confidence duplicate pairs. Then, a \emph{visual verification} step authorizes or rejects each merge.

To avoid overwhelming the VLM with spurious proposals, we apply a two-gate filtering mechanism to each pair of tracks $(T_i, T_j)$, including tentative ones, with observation counts $m_i, m_j$ (the number of successfully associated detections for each track) and 3D bounding boxes $B_i, B_j$. A \emph{geometric gate} first discards any pair with $\min(m_i, m_j) < 2$ or $\mathrm{IoU}_{3\mathrm{D}}(B_i, B_j) < \tau_{\text{iou}}$,
pruning noisy single-observation detections. Surviving pairs pass through a \emph{semantic gate} that scores each pair by the cosine similarity $s_{ij}$ between the SBERT~\cite{reimers_sentence-bert_2019} embeddings of their consensus labels. We use SBERT rather than CLIP here, as CLIP gives uniformly high similarity even for distinct labels, making threshold filtering unreliable.

Pairs with $s_{ij} \ge \tau_{\text{sem}}$ enter a candidate buffer that decouples tracking from verification.
For each buffered pair, we track the peak alignment via a running maximum $\hat{s}_{ij} \leftarrow \max(\hat{s}_{ij},\, s_{ij}^{(t)})$, capturing the best-observed compatibility as new evidence arrives.
The buffer is flushed periodically, ranking candidates by $\hat{s}_{ij}$ and dispatching the top-$K_{\text{critic}}$ to the verifier.
At the end of the sequence, all remaining valid pairs are flushed to ensure complete coverage.
When merging two tentative tracks results in a combined observation count that exceeds $N_{\text{conf}}$, the merged track is automatically promoted to a confirmed node.

For each selected pair, we first keep high-visibility views of each track, measured by visible mask area. If both tracks are visible in the same frame, we  choose the shared frame where both are most visible. Otherwise, we choose two high-visibility views with similar camera direction, camera position, object depth, and temporal proximity.
To ground VLM attention without obscuring texture details, we employ Set-of-Mark (SoM) prompting~\cite{yang2023setofmark}, overlaying high-contrast SAM mask boundaries and unique identifiers directly onto the image.
The model receives these marked images alongside a context tuple $\mathcal{C} = \{ \text{ID}_{i,j}, \text{Label}_{i,j}, \hat{s}_{ij}, \text{geometry cues} \}$ that supplies geometric cues to reduce hallucinations from ambiguous masks, and returns a structured decision $\mathcal{D} = \{ a \in \{\textsc{Merge}, \textsc{Keep}\}, r \in \text{String} \}$. When the verifier returns $a=\textsc{Merge}$, the Critic unifies point clouds, features, and label histograms of both tracks in the scene graph.

\paragraph{Description agent. } 
While the tracker maintains object identities and coarse text labels, downstream tasks such as visual grounding require deeper semantic understanding. We introduce a Description Agent that runs asynchronously in a dedicated worker thread to enrich each scene graph node with fine-grained visual attributes, such as material, color, texture and physical state, which are critical for disambiguating instances of the same class.

Since a single frame typically depicts multiple objects, a small set of well-chosen frames can cover all targets while amortizing VLM cost and preserving scene context that individual crops would lack.
We therefore introduce a multi-target frame scheduler that buffers 2D bounding boxes over a tumbling window of $W_\text{desc}$ frames and identifies targets that lack a description or whose best visible area substantially exceeds their last-described area.
It then greedily selects up to $K_\text{desc}$ frames that maximize target coverage.
A target $T$ can be covered in frame $f$ only if its normalized screen area $\hat{A}_{T,f}$ is within a fraction $\tau_\text{view}$ of its best view in the window. Each candidate frame is scored by
\begin{equation}
  U(f) = \sum_{T \in \mathcal{T}_f} w_T \cdot \ln\!\bigl(1 + \gamma\,\hat{A}_{T,f}\bigr),
  \label{eq:utility}
\end{equation}
where $\mathcal{T}_f$ is the set of targets coverable in frame $f$, $w_T$ is a per-target weight that down-weights background classes and $\gamma$ controls the area sensitivity. The greedy loop selects the highest-scoring frame, marks its targets as covered, and repeats until the budget is exhausted.

Selected frames are also rendered with SoM overlays and submitted in a single multi-image VLM call, along with each target's current label, description, and attribute estimates as context.
For each marked object, the model returns a refined label (\eg, ``winged armchair'' rather than ``chair'') and visual attributes (material, color, state, texture).
Attributes accumulate in a per-node histogram mirroring the confidence-weighted voting of \cref{eq:label_vote}, so repeated observations reinforce consistent evidence.

\section{Experiments}
\label{sec:experiments}

We evaluate on open-vocabulary 3D semantic segmentation and 3D visual grounding, then analyze component contributions through ablations.

\paragraph{Datasets.}
For semantic segmentation, we evaluate on Replica~\cite{straub2019replica} (eight photorealistic scenes: \texttt{room0}--\texttt{room2}, \texttt{office0}--\texttt{office4}) and ScanNet~\cite{dai_scannet_2017} (eight real-world RGB-D sequences: \texttt{0011\_00}, \texttt{0030\_00}, \texttt{0046\_00}, \texttt{0086\_00}, \texttt{0222\_00}, \texttt{0378\_00}, \texttt{0389\_00}, \texttt{0435\_00}).
For visual grounding, we use three referring expression benchmarks with ScanNet: ScanRefer~\cite{chen_scanrefer_2020} (natural-language descriptions), Sr3D+~\cite{achlioptas2020referit3d} (template-generated), and Nr3D~\cite{achlioptas2020referit3d} (free-form human-written).
For both tasks, we follow the evaluation protocol of BBQ~\cite{linok_bare_2025}; the eight ScanNet scenes are chosen to match their grounding setup. We adopt the same frame sampling stride of 5 for Replica and 10 for ScanNet.

\paragraph{Metrics.}
For segmentation, we report mean per-class accuracy (mAcc), mean intersection-over-union (mIoU), and frequency-weighted IoU (f-mIoU), following the open-vocabulary protocol of BBQ~\cite{linok_bare_2025} with EVA02-CLIP~\cite{fang2023eva02} features. Per-point predictions are matched to ground truth via nearest-neighbor association in 3D. Generic categories (\textit{other}, \textit{otherfurniture}) are excluded. For grounding, we use GPT-4o~\cite{openai2024gpt4o} and report accuracy at IoU thresholds A@0.25 and A@0.5 on ScanRefer, and at A@0.1 and A@0.25 on Sr3D+ and Nr3D. The full evaluator prompt and an ablation over evaluator models are provided in the supplementary.

\paragraph{Implementation.}
For our CLIP embeddings we use the EVA02-CLIP~\cite{fang2023eva02} variant. Qwen3-VL-2B-Instruct~\cite{bai_qwen3vl_2025} runs locally; for the Agents we use GPT-5-mini~\cite{openai_gpt5_2025}, called via its cloud API. Experiments use a single workstation (AMD Ryzen 7 9700X, NVIDIA RTX 5090, 64\,GB RAM). We set the voxel backend to a resolution of 4\,cm with an adaptive area ratio $\rho_{\text{area}}{=}0.5$. Association balances geometry and feature similarity ($\lambda_{\text{geo}}{=}0.8$, $\lambda_{\text{feat}}{=}0.2$) at a threshold $\tau_{\text{assoc}}{=}0.4$; tentative tracks are promoted after $N_{\text{conf}}{=}6$ successful associations. The semantic gate is set to $\tau_{\text{sem}}{=}0.8$ and the IoU gate to $\tau_{\text{iou}}{=}0.01$, with utility scale $\gamma{=}100$. The Critic Agent flushes up to $K_{\text{critic}}{=}20$ candidate pairs every $W_{\text{critic}}{=}50$ frames. The Description Agent operates over tumbling windows of $W_{\text{desc}}{=}30$ frames, selecting up to $K_{\text{desc}}{=}3$ frames per window. The remaining hyperparameters are listed in the supplementary.
\section{Results}
\label{sec:results}
We evaluate our method on two tasks: open-vocabulary 3D semantic segmentation and 3D visual grounding. We then analyze the contribution of individual components through ablations.


\subsection{Open-Vocabulary 3D Semantic Segmentation}

\begin{wrapfigure}{r}{0.45\linewidth}
    \centering
    \vspace{-1.5em}
    \includegraphics[width=\linewidth]{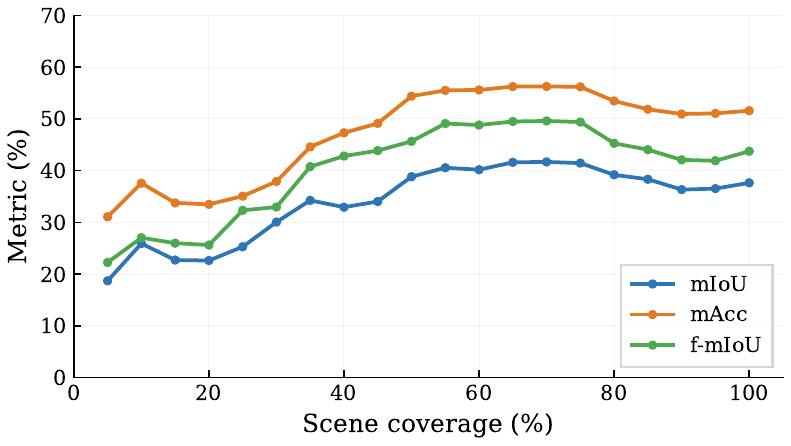}
    \caption{\textbf{Incremental Segmentation.} On Room0 (Replica), metrics converge steadily as frames are processed.}
    \label{fig:semseg_time}
    \vspace{-2em}
\end{wrapfigure}

\cref{tab:semseg} compares our method with prior open-vocabulary 3D methods on Replica and ScanNet. On Replica, our incremental method outperforms all prior methods on every metric (0.58 mAcc, 0.37 mIoU, 0.61 f-mIoU), with the largest gain on f-mIoU (+0.13 over BBQ-CLIP).
On ScanNet, we obtain 0.75 mAcc, 0.44 mIoU, and 0.46 f-mIoU, achieving the highest mIoU and f-mIoU while remaining competitive with OpenVox (0.76) on mAcc.
We attribute these gains to two factors.
First, the voxel-consistency association with CLIP text-label embeddings produces more stable track identities than the pairwise visual similarity used by prior methods.
Second, the text-guided feature selection in \cref{eq:text_guided} ensures that each track is represented by the view most consistent with its consensus label, producing cleaner per-point semantic assignments.
\cref{fig:semseg_time} illustrates this effect over time: Metrics improve steadily as additional frames are processed, demonstrating that the incremental pipeline converges to high-quality predictions without requiring a complete reconstruction.

\begin{table}[tb]
  \centering
  \small
  \setlength{\tabcolsep}{4pt}
  \caption{\textbf{3D Semantic Segmentation.} Comparison with prior open-vocabulary methods on Replica and ScanNet.}
  \label{tab:semseg}
  \begin{tabular}{@{}lccc@{\hskip 8pt}ccc@{}}
    \toprule
    & \multicolumn{3}{c}{\textbf{Replica}} & \multicolumn{3}{c}{\textbf{ScanNet}} \\
    \cmidrule(lr){2-4} \cmidrule(lr){5-7}
    \textbf{Method} & \textbf{mAcc} & \textbf{mIoU} & \textbf{f-mIoU} & \textbf{mAcc} & \textbf{mIoU} &\textbf{f-mIoU} \\
    \midrule
    ConceptFusion~\cite{jatavallabhula2023conceptfusion} & 0.29 & 0.11 & 0.14 & 0.49 & 0.26 & 0.31 \\
    OpenMask3D~\cite{takmaz2023openmask3db} & -- & -- & -- & 0.34 & 0.18 & 0.20 \\
    ConceptGraphs~\cite{gu2024conceptgraphs} & 0.36 & 0.18 & 0.15 & 0.52 & 0.26 & 0.29 \\
    BBQ-CLIP~\cite{linok_bare_2025} & 0.38 & 0.27 & \cellcolor{second}0.48 & 0.56 & 0.34 & 0.36 \\
    Octree-Graph~\cite{wang2025openvocabulary} & \cellcolor{second}0.51 & \cellcolor{second}0.34 & 0.34 & 0.71 & 0.40 & 0.37 \\
    OpenVox~\cite{yinan_openvox_2025} & \cellcolor{second}0.51 & 0.26 & 0.40 & \cellcolor{best}0.76 & \cellcolor{second}0.43 & \cellcolor{second}0.39 \\
    \midrule
    \textbf{ThinkGraphs} & \cellcolor{best}\textbf{0.58} & \cellcolor{best}\textbf{0.37} & \cellcolor{best}\textbf{0.61} & \cellcolor{second}0.75 & \cellcolor{best}\textbf{0.44} & \cellcolor{best}\textbf{0.46} \\
    \bottomrule
  \end{tabular}
\end{table}

\subsection{3D Visual Grounding}
We evaluate on three referring expression benchmarks of increasing difficulty: the template-based Sr3D+, the free-form Nr3D, and the natural-language ScanRefer. To handle the complexity of free-form and natural-language evaluation, we utilize GPT-4o~\cite{openai2024gpt4o} as our automated evaluator.

\paragraph{Sr3D+ and Nr3D.}
On both referral benchmarks (\cref{tab:sr3d_nr3d_overall}), our method substantially outperforms all baselines, improving over Open3DSG~\cite{koch2024open3dsg} by 15.3 A@0.25 on Sr3D+ and 18.8 on Nr3D. Performance remains consistent across difficulty and viewpoint splits (see supplementary for per-split breakdowns).

\paragraph{ScanRefer.}
On the ScanRefer benchmark~(\cref{tab:scanrefer}), our method achieves 52.9\% A@0.25 and 37.6\% A@0.5, outperforming Open3DSG~\cite{koch2024open3dsg} by 18.3 and 6.6 absolute points respectively. This reflects the combined effect of our core components: our backend provides stable track identities and high-quality representative visual embeddings, the Critic Agent reduces node fragmentation so that queries match the correct consolidated object, and the Description Agent provides fine-grained attributes that help disambiguate same-class instances.

\begin{table*}[t]
\centering

\begin{minipage}[t]{0.55\linewidth}
    \centering
    \caption{\textbf{3D Visual Grounding.} Overall accuracy on Sr3D+ and Nr3D.}
    \label{tab:sr3d_nr3d_overall}
    \begin{tabular}{@{}lcccc@{}}
        \toprule
        & \multicolumn{2}{c}{\textbf{Sr3D+}} 
        & \multicolumn{2}{c}{\textbf{Nr3D}} \\
        \cmidrule(lr){2-3} \cmidrule(lr){4-5}
        \textbf{Method} 
        & A@0.1 & A@0.25 
        & A@0.1 & A@0.25 \\
        \midrule
        OpenFusion~\cite{yamazaki_openfusion_2024} & 12.6 & 2.4 & 10.7 & 1.4 \\
        BBQ-CLIP~\cite{linok_bare_2025} & 14.4 & 8.8 & 15.3 & 9.4 \\
        ConceptGraphs~\cite{gu2024conceptgraphs} & 13.3 & 6.2 & 16.0 & 7.2 \\
        BBQ~\cite{linok_bare_2025} & \cellcolor{second}34.2 & 22.7 & \cellcolor{second}28.3 & 19.0 \\
        Open3DSG~\cite{koch2024open3dsg} & 28.9 & \cellcolor{second}27.7 & 25.9 & \cellcolor{second}23.0 \\
        \midrule
        \textbf{ThinkGraphs} 
        & \cellcolor{best}\textbf{48.6} 
        & \cellcolor{best}\textbf{43.0} 
        & \cellcolor{best}\textbf{47.4} 
        & \cellcolor{best}\textbf{41.8} \\
        \bottomrule
    \end{tabular}
\end{minipage}
\hfill
\begin{minipage}[t]{0.40\linewidth}
    \centering
    \caption{\textbf{3D Visual grounding.} Overall accuracy on ScanRefer.}
    \label{tab:scanrefer}
    \begin{tabular}{@{}lcc@{}}
        \toprule
        & \multicolumn{2}{c}{\textbf{ScanRefer}} \\
        \cmidrule(lr){2-3}
        \textbf{Method} & A@0.25 & A@0.5 \\
        \midrule
        LERF~\cite{kerr_lerf_2023}          & 4.4  & 0.3 \\
        OpenScene~\cite{peng_openscene_2023}     & 13.0 & 5.1 \\
        LLM-Grounder~\cite{yang_llmgrounder_2024}  & 17.1 & 5.3 \\
        BBQ~\cite{linok_bare_2025} & 19.4 & 11.6 \\
        Open3DSG~\cite{koch2024open3dsg} & \cellcolor{second}34.6 & \cellcolor{second}31.0 \\
        \midrule
        \textbf{ThinkGraphs} 
        & \cellcolor{best}\textbf{52.9} 
        & \cellcolor{best}\textbf{37.6} \\
        \bottomrule
    \end{tabular}
\end{minipage}

\end{table*}

\subsection{Ablation Studies}
To isolate individual component contributions, we conduct ablations on the eight Replica scenes.

\paragraph{Pipeline design choices.}
\cref{tab:semseg_ablation} ablates the backend pipeline on Replica.
Switching from RAM++~\cite{huang_openset_2025} to Qwen3-VL~\cite{bai_qwen3vl_2025} as the frontend tagger yields the largest mAcc gain (+0.09), as context-aware prompts produce fewer missed detections.
Text-guided embedding selection drives the largest mIoU gain (+0.09): Naively selecting the largest-area crop as the representative embedding, as done \eg in BBQ~\cite{linok_bare_2025}, is sensitive to foreground clutter.
Scoring against the consensus label produces cleaner views.
The adaptive area gate adds a further 0.06 mAcc by filtering low-quality observations, while replacing SBERT with CLIP text-label embeddings provides a smaller benefit but unifies the visual and textual feature space.

\begin{table}[tb]
  \centering
  \small
  \caption{\textbf{Pipeline ablation.} Component contributions on Replica; after replacing the RAM++ baseline with Qwen, components are added incrementally.}
  \label{tab:semseg_ablation}
  \begin{tabular}{@{}l ccc@{}}
    \toprule
    \textbf{Configuration} & \textbf{mAcc} & \textbf{mIoU} & \textbf{f-mIoU} \\
    \midrule
    RAM++ (baseline frontend)          & 0.37 & 0.20 & 0.34 \\
    Qwen (our frontend)              & 0.46 & 0.25 & 0.43 \\
    \midrule
    \quad + CLIP-text labels         & 0.47 & 0.26 & 0.44 \\
    \quad + Text-guided selection    & 0.52 & 0.35 & 0.55 \\
    \quad + Area filter              & \textbf{0.58} & \textbf{0.37} & \textbf{0.61} \\
    \bottomrule
  \end{tabular}
\end{table}

\begin{figure}[tb]
    \centering
    \includegraphics[width=\linewidth]{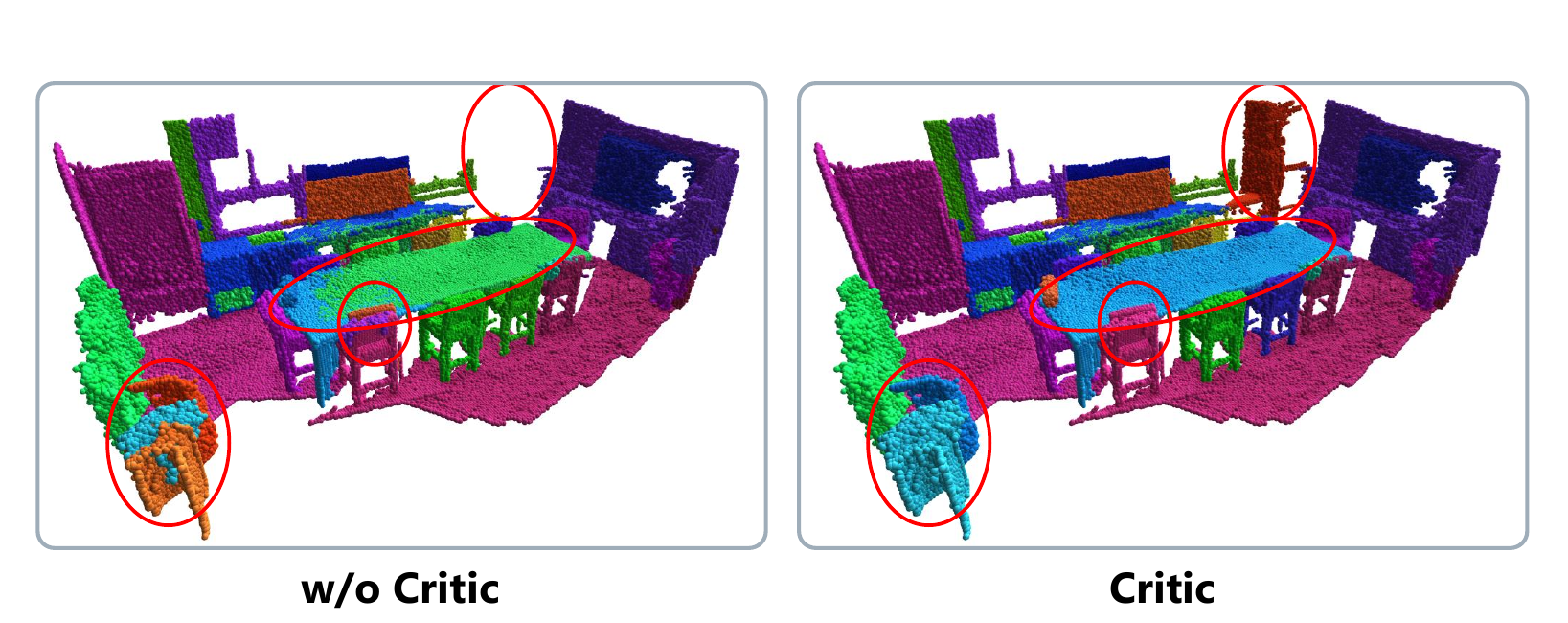}
    \caption{\textbf{Effect of the Critic Agent.} Without the Critic (left),
    association drift fragments objects into duplicate object tracks (red circles).
    Semantic loop closure (right) merges them into consistent identities.
    Points are colored by instance.}
    \label{fig:critic_no_critic}
\end{figure}

\paragraph{Agent contributions to grounding.}
\cref{tab:nr3d_ablation} incrementally adds the two asynchronous agents to isolate their effect on Nr3D grounding accuracy.
The base system includes the full backend with spatial relations, but no VLM agents. Adding the Critic Agent alone lifts overall A@0.25 from 38.5\% to 39.3\%, with gains concentrated on easy and view-independent splits where fragmented object tracks are the primary bottleneck (\cref{fig:critic_no_critic}).
The Description Agent has a larger effect on hard referrals (+3.5 A@0.25), where fine-grained attributes are needed to disambiguate same-class instances.
Combining both agents yields the best results in all splits (41.8\% A@0.25 overall, +7.1 on hard), confirming that the two agents address complementary failure modes. An ablation isolating the architecture from the frontend VLM (a lighter RAM++ tagger) is provided in the supplementary, showing the gains are not merely due to a stronger tagger.

\begin{table}[tb]
    \caption{\textbf{Agent ablation.} Effect of the Critic and Description agents on Nr3D; Base is the online mapping pipeline without the VLM agents.}
    \label{tab:nr3d_ablation}
    \resizebox{\linewidth}{!}{%
    \begin{tabular}{@{}lcccccccccc@{}}
    \toprule
    & \multicolumn{2}{c}{\textbf{Overall}} & \multicolumn{2}{c}{\textbf{Easy}} & \multicolumn{2}{c}{\textbf{Hard}} & \multicolumn{2}{c}{\textbf{View Dep.}} & \multicolumn{2}{c}{\textbf{View Indep.}} \\
    \cmidrule(lr){2-3} \cmidrule(lr){4-5} \cmidrule(lr){6-7} \cmidrule(lr){8-9} \cmidrule(lr){10-11}
    \textbf{Method}
    & A@0.1 & A@0.25 & A@0.1 & A@0.25 
    & A@0.1 & A@0.25 & A@0.1 & A@0.25 
    & A@0.1 & A@0.25 \\
    \midrule
    Base 
    & 43.1 & 38.5 
    & 49.2 & 46.2 
    & 27.4 & 18.8 
    & 43.0 & 37.6 
    & 43.1 & 38.8 \\
    
    Base + Critic
    & 45.1 & 39.3
    & 52.2 & 47.4
    & 26.9 & 18.8
    & 43.0 & 38.2
    & 45.7 & 39.7 \\
    
    Base + Description
    & 44.5 & 40.5
    & 50.0 & 47.6
    & 30.5 & 22.3
    & 44.8 & 40.6
    & 44.4 & 40.4 \\
    \midrule
    Base + Critic + Description
    & \textbf{47.4} & \textbf{41.8}
    & \textbf{53.0} & \textbf{48.0}
    & \textbf{33.0} & \textbf{25.9}
    & \textbf{49.1} & \textbf{43.0}
    & \textbf{46.8} & \textbf{41.4} \\
    
    \bottomrule
    \end{tabular}
    }
\end{table}


\paragraph{Runtime and cost.}
The online mapping path averages 1.45\,s per keyframe, with the Critic and Description
agents running off the critical path so they never block the online mapping loop. The
multi-target scheduler keeps VLM usage low, about 26 calls per scene on average,  more than two
orders of magnitude fewer than naive per-object-per-keyframe prompting ($\sim$9{,}300 on
a typical scene). The primary runtime bottleneck is the frontend, not the agents. A full
analysis is provided in the supplementary.

\subsection{Limitations}
Our framework has two main limitations. Firstly, the front-end models (Qwen3-VL + Grounded-SAM) prevent real-time operation. Replacing these models with more efficient alternatives would be the most direct way of overcoming this issue. Secondly, asynchronous agents inevitably lag behind the mapping frontier, meaning that objects discovered during fast exploration may remain unenriched for several frames. This limits the quality of grounding early in a sequence.
\section{Conclusion}
\label{sec:conclusion}

We present ThinkGraphs, an asynchronous architecture for open-vocabulary 3D scene graphs that decouples lightweight online mapping from heavyweight VLM reasoning. By running semantic refinement in the background, the scene graph remains queryable during exploration and grows in richness over time. A probabilistic voxel backbone with text-guided embedding selection provides stable object identities, while a Critic Agent corrects fragmentation through semantic loop closure, and a Description Agent attaches fine-grained attributes via multi-target frame scheduling.
Our method outperforms all batch-based open-vocabulary scene-graph methods on semantic segmentation on Replica and ScanNet, and surpasses the prior state-of-the-art across three visual grounding benchmarks (Sr3D+, Nr3D, ScanRefer) by 15.3 to 18.8 A@0.25, showing that an incremental architecture can match and exceed batch-based alternatives. Future work could extend this asynchronous agent design with additional specialized agents, for example for richer relation prediction, higher-level scene reasoning, or task-oriented graph refinement, while also exploring its use in other structured representations that require both responsiveness and semantic depth.

\section*{Acknowledgements}
The authors thank the International Max Planck Research School for Intelligent Systems (IMPRS-IS) for supporting Deniz Bickici and Michael Pabst.
This work was supported by the Alexander von Humboldt Foundation funded by the German Federal Ministry of Research, Technology and Space, and by Deutsche Forschungsgemeinschaft (DFG) under Germany's Excellence Strategy EXC 2120/2 (390831618).


%
%
\bibliographystyle{splncs04}
\bibliography{main}

\clearpage
\appendix

\begin{center}
  {\Large\bfseries Think While You Map:\\Asynchronous Vision-Language Agents for Incremental 3D Scene Graphs \par}
  \vspace{0.8em}
  {\large Supplementary Material \par}
  \vspace{0.4em}\end{center}
\FloatBarrier

\section{Per-Split Grounding Results}
\label{sec:full_results}

\cref{tab:nr3d_grounding,tab:sr3d_grounding,tab:scanrefer_splits} expand the overall grounding numbers from the main paper (Tabs.~2 and~3) into per-split breakdowns.
On Nr3D and Sr3D+~\cite{achlioptas2020referit3d}
(\cref{tab:nr3d_grounding,tab:sr3d_grounding}), ThinkGraphs outperforms BBQ~\cite{linok_bare_2025} across nearly all splits. The one exception is Sr3D+'s hard split at A@0.1, where BBQ has a slight advantage (33.3 vs. 32.0). However, ThinkGraphs still leads at A@0.25 (32.0 vs. 22.7).
For ScanRefer~\cite{chen_scanrefer_2020} (\cref{tab:scanrefer_splits}), we analyze performance by split. Our approach reaches 71.8\% A@0.25 on the unique split. Performance remains also strong on the more challenging multiple split, achieving 49.0\% A@0.25, indicating that the attributes of the Description Agent effectively disambiguate instances belonging to the same class.

To isolate the contribution of our architecture, we replace the Qwen tagger with the more lightweight RAM++~\cite{huang_openset_2025} tagger and rerun the full system on Nr3D. Even with this weaker frontend, ThinkGraphs reaches 33.6 A@0.25, still exceeding Open3DSG (23.0) and BBQ (19.0). The 18.8 A@0.25 overall gain over Open3DSG thus splits into 10.6 from the architecture and 8.2 from the Qwen3-VL tagger, confirming that the improvement does not stem solely from a stronger tagger.


\begin{table}[tb]
\caption{\textbf{Nr3D Grounding.} Per-split accuracy on Nr3D. \textit{ThinkGraphs (RAM++)} replaces our Qwen3-VL tagger with the lighter RAM++ tagger.}
\label{tab:nr3d_grounding}
\centering
\resizebox{\linewidth}{!}{%
\begin{tabular}{@{}lcccccccccc@{}}
\toprule
& \multicolumn{2}{c}{\textbf{Overall}} & \multicolumn{2}{c}{\textbf{Easy}} & \multicolumn{2}{c}{\textbf{Hard}} & \multicolumn{2}{c}{\textbf{View Dep.}} & \multicolumn{2}{c}{\textbf{View Indep.}} \\
\cmidrule(lr){2-3} \cmidrule(lr){4-5} \cmidrule(lr){6-7} \cmidrule(lr){8-9} \cmidrule(lr){10-11}

\textbf{Method}
& A@0.1 & A@0.25 & A@0.1 & A@0.25 
& A@0.1 & A@0.25 & A@0.1 & A@0.25 
& A@0.1 & A@0.25 \\
\midrule
OpenFusion~\cite{yamazaki_openfusion_2024}   & 10.7 & 1.4 & 12.9 & 1.4 & 5.1 & 1.5 & 8.5 & 0.0 & 11.4 & 1.9 \\
BBQ-CLIP~\cite{linok_bare_2025}     & 15.3 & 9.4 & 18.1 & 11.0 & 8.1 & 5.6 & 8.1 & 6.1 & 17.2 & 10.5 \\
ConceptGraphs~\cite{gu2024conceptgraphs} & 16.0 & 7.2 & 18.7 & 9.2 & 9.1 & 2.0 & 12.7 & 4.2 & 17.0 & 8.1 \\
BBQ~\cite{linok_bare_2025} & 
\cellcolor{second}28.3 & 19.0 
& \cellcolor{second}30.5 & 21.3 
& 22.8 & 13.2 
& 23.6 & 18.2 
& \cellcolor{second}29.8 & 19.3 \\

Open3DSG~\cite{koch2024open3dsg} &
25.9 & \cellcolor{second}23.0
& 26.5 & \cellcolor{second}25.5
& \cellcolor{second}24.4 & \cellcolor{second}16.8
& \cellcolor{second}26.1 & \cellcolor{second}23.6
& 25.8 & \cellcolor{second}22.8 \\

\midrule
\textbf{ThinkGraphs} &
\cellcolor{best}\textbf{47.4} & \cellcolor{best}\textbf{41.8} 
& \cellcolor{best}\textbf{53.0} & \cellcolor{best}\textbf{48.0} 
& \cellcolor{best}\textbf{33.0} & \cellcolor{best}\textbf{25.9} 
& \cellcolor{best}\textbf{49.1} & \cellcolor{best}\textbf{43.0} 
& \cellcolor{best}\textbf{46.8} & \cellcolor{best}\textbf{41.4} \\
ThinkGraphs (RAM++) &
39.3 & 33.6
& 45.8 & 41.6
& 22.8 & 13.2
& 37.6 & 30.9
& 39.9 & 34.5 \\
\bottomrule
\end{tabular}
}
\end{table}


\begin{table}[tb]
\caption{\textbf{Sr3D+ Grounding.} Per-split accuracy on Sr3D+}
\label{tab:sr3d_grounding}
\centering
\resizebox{\linewidth}{!}{%
\begin{tabular}{@{}lcccccccccc@{}}
\toprule
& \multicolumn{2}{c}{\textbf{Overall}} & \multicolumn{2}{c}{\textbf{Easy}} & \multicolumn{2}{c}{\textbf{Hard}} & \multicolumn{2}{c}{\textbf{View Dep.}} & \multicolumn{2}{c}{\textbf{View Indep.}} \\
\cmidrule(lr){2-3} \cmidrule(lr){4-5} \cmidrule(lr){6-7} \cmidrule(lr){8-9} \cmidrule(lr){10-11}
\textbf{Method}
& A@0.1 & A@0.25 & A@0.1 & A@0.25 
& A@0.1 & A@0.25 & A@0.1 & A@0.25 
& A@0.1 & A@0.25 \\
\midrule
OpenFusion~\cite{yamazaki_openfusion_2024}   
& 12.6 & 2.4 & 14.0 & 2.4 & 1.3 & 1.3 & 3.8 & 2.5 & 13.7 & 2.4 \\

BBQ-CLIP~\cite{linok_bare_2025}     
& 14.4 & 8.8 & 15.4 & 9.0 & 6.7 & 6.7 & 11.4 & 5.1 & 14.4 & 8.8 \\

ConceptGraphs~\cite{gu2024conceptgraphs} 
& 13.3 & 6.2 & 13.0 & 6.8 & 16.0 & 1.3 & 15.2 & 5.1 & 13.1 & 6.4 \\

BBQ~\cite{linok_bare_2025} 
& \cellcolor{second}34.2 & 22.7 
& \cellcolor{second}34.3 & 22.7 
& \cellcolor{best}\textbf{33.3}& \cellcolor{second}22.7 
& \cellcolor{second}32.9 & 20.3 
& \cellcolor{second}34.4 & 23.0 \\

Open3DSG~\cite{koch2024open3dsg}
& 28.9 & \cellcolor{second}27.7
& 29.9 & \cellcolor{second}28.8
& 21.3 & 18.7
& 29.1 & \cellcolor{second}29.1
& 28.9 & \cellcolor{second}27.5 \\

\midrule
\textbf{ThinkGraphs} 
& \cellcolor{best}\textbf{48.6} & \cellcolor{best}\textbf{43.0} 
& \cellcolor{best}\textbf{50.7} & \cellcolor{best}\textbf{44.4} 
& \cellcolor{second}32.0 & \cellcolor{best}\textbf{32.0} 
& \cellcolor{best}\textbf{53.2} & \cellcolor{best}\textbf{48.1} 
& \cellcolor{best}\textbf{47.9} & \cellcolor{best}\textbf{42.3} \\
\bottomrule
\end{tabular}
}
\end{table}

\begin{table}[tb]
\centering
\caption{\textbf{ScanRefer Grounding.} Per-split accuracy on ScanRefer.}
\label{tab:scanrefer_splits}
\begin{tabular}{lcccccc}
\toprule
& \multicolumn{2}{c}{\textbf{Overall}} & \multicolumn{2}{c}{\textbf{Unique}} & \multicolumn{2}{c}{\textbf{Multiple}} \\
\cmidrule(lr){2-3} \cmidrule(lr){4-5} \cmidrule(lr){6-7}
\textbf{Method} & A@0.25 & A@0.5 & A@0.25 & A@0.5 & A@0.25 & A@0.5 \\
\midrule
\textbf{ThinkGraphs} & \textbf{52.9} & \textbf{37.6} & \textbf{71.8} & \textbf{52.4} & \textbf{49.0} & \textbf{34.6} \\
\bottomrule
\end{tabular}
\end{table}

\section{Grounding Model Ablation}
Our main-paper results use GPT-4o~\cite{openai2024gpt4o} as the grounding evaluator, following the protocol of BBQ~\cite{linok_bare_2025}. To test sensitivity to the evaluator, we re-evaluate the same ThinkGraphs scene graphs with GPT-5.4~\cite{openai_gpt54_2026} and Llama-4 Maverick~\cite{meta2025llama4} (\cref{tab:nr3d_model_ablation}) on Nr3D. Overall accuracy stays within a narrow band around GPT-4o, ranging from a 1.9 A@0.25 drop with Llama-4 Maverick to a 3.6 gain with GPT-5.4, confirming that our improvements do not depend on a particular evaluator.

\begin{table}[tb]
\caption{\textbf{Evaluator Ablation.} Effect of the grounding evaluator on Nr3D; scene graphs are fixed (produced by ThinkGraphs), only the evaluator model varies.}
\label{tab:nr3d_model_ablation}
\resizebox{\linewidth}{!}{%
\begin{tabular}{@{}lcccccccccc@{}}
\toprule
& \multicolumn{2}{c}{\textbf{Overall}} & \multicolumn{2}{c}{\textbf{Easy}} & \multicolumn{2}{c}{\textbf{Hard}} & \multicolumn{2}{c}{\textbf{View Dep.}} & \multicolumn{2}{c}{\textbf{View Indep.}} \\
\cmidrule(lr){2-3} \cmidrule(lr){4-5} \cmidrule(lr){6-7} \cmidrule(lr){8-9} \cmidrule(lr){10-11}

\textbf{Evaluator}
& A@0.1 & A@0.25 & A@0.1 & A@0.25
& A@0.1 & A@0.25 & A@0.1 & A@0.25
& A@0.1 & A@0.25 \\
\midrule

GPT-4o~\cite{openai2024gpt4o}
& 47.4 & 41.8
& 53.0 & 48.0
& 33.0 & 25.9
& \textbf{49.1} & \textbf{43.0}
& 46.8 & 41.4 \\

GPT-5.4~\cite{openai_gpt54_2026}
& \textbf{51.5} & \textbf{45.4}
& \textbf{56.0} & \textbf{51.6}
& \textbf{40.1} & \textbf{29.4}
& 43.0 & 38.8
& \textbf{54.1} & \textbf{47.4} \\

Llama-4 Maverick~\cite{meta2025llama4}
& 45.8 & 39.9
& 49.4 & 44.4
& 36.5 & 28.4
& 43.6 & 38.2
& 46.4 & 40.4 \\

\bottomrule
\end{tabular}
}
\end{table}

\section{Agent Behavior Analysis}
\label{sec:analysis}

We provide an additional analysis of the agents' behavior.

\begin{figure}[tb]
    \centering
    \includegraphics[width=1\linewidth]{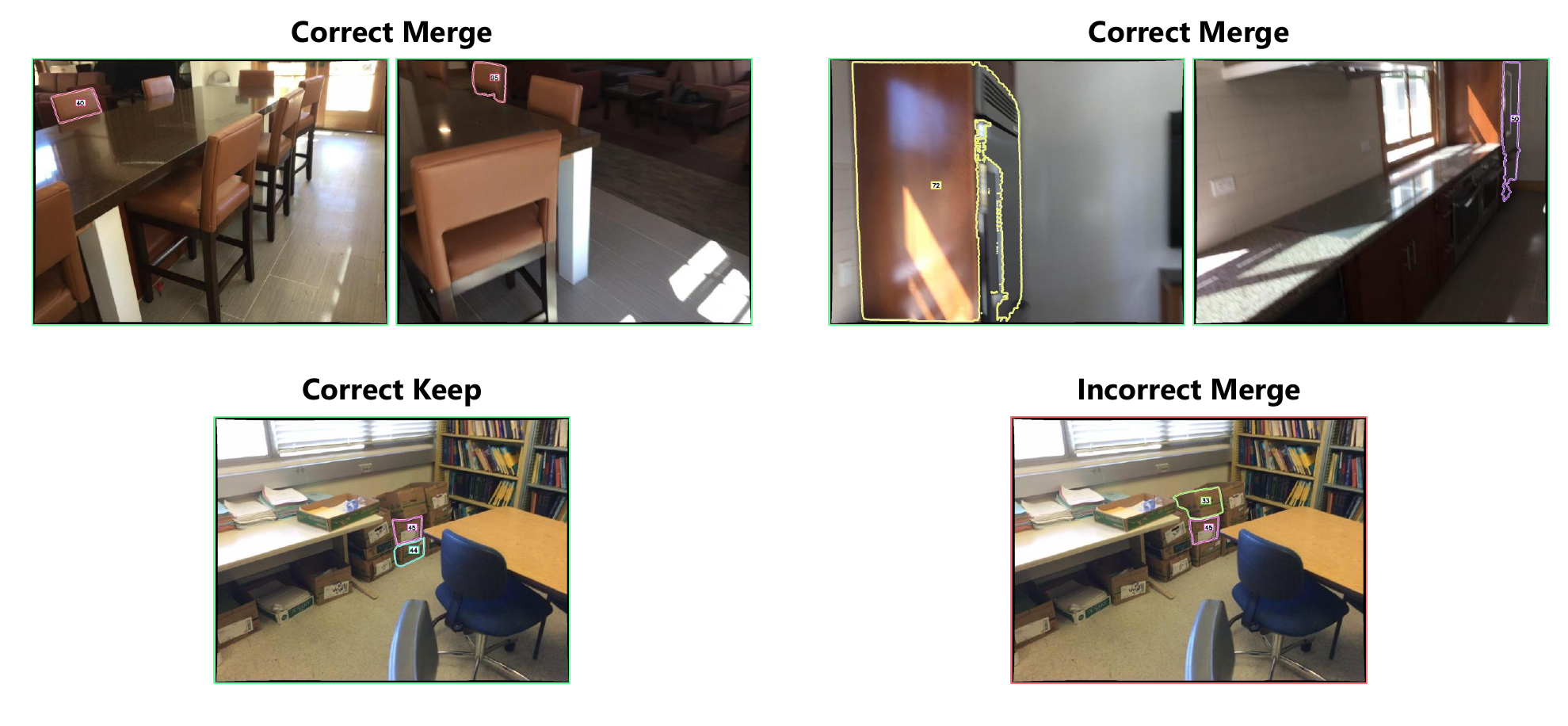}
    \caption{\textbf{Critic Agent Decisions.} Examples on ScanNet; green borders mark correct decisions, red borders errors.}
    \label{fig:critic_examples}
\end{figure}

\paragraph{Critic Agent behavior.}
\cref{fig:critic_examples} shows representative decisions of the Critic Agent. In the first two cases, the agent correctly identifies duplicate tracks caused by viewpoint changes, such as a leather chair observed from two angles or a partially visible appliance fragmented across frames. In both cases, the agent reasons about spatial offset, containment, and visual consistency to authorize the merge. These are precisely the cases where greedy CLIP-similarity and 3D-IoU association fails: the viewpoint change lowers both embedding similarity and geometric overlap, so a pairwise matcher leaves the object fragmented, whereas the Critic recovers the merge from visual evidence. The third case illustrates a correct rejection. Two visually similar boxes stacked vertically are kept as distinct instances due to their low containment (7.1\%) and vertical separation (0.23\,m). Conversely, this is a case where IoU- or containment-based merging would erroneously fuse two distinct same-class instances; the Critic keeps them separate. The fourth case illustrates a failure mode in which the axis-aligned bounding box of a larger object overestimates its spatial extent. This causes a smaller, nearby box to appear highly contained and be incorrectly merged as a fragment.

To quantify these behaviors across scenes,~\cref{tab:critic_behavior}
categorizes all Critic decisions into three types: \emph{deduplications}, which merge two confirmed tracks referring to the same object and remove a redundant node; \emph{absorptions}, which fold an unconfirmed fragment into an existing confirmed track, improving its geometry without changing the node count; and \emph{promotions}, which merge two unconfirmed fragments whose combined observations exceed~$N_{\text{conf}}$ and recover an otherwise lost object. Across the eight scenes, 60 out of 144 calls (41.7\%) are accepted. Absorptions dominate (38 of 60), followed by 17~deduplications and 5~promotions. Because promotions partially offset deduplications in the total confirmed count, the Critic's full contribution is best measured through the downstream grounding ablation in Tab.~5 of the main paper.

\begin{table}[tb]
\centering
\caption{\textbf{Critic Decisions by Type.} Counts across the eight ScanNet benchmark scenes. \emph{Dedup.}\ merges two confirmed tracks, removing a redundant node. \emph{Absorb.}\ folds an unconfirmed fragment into a confirmed track, improving its geometry. \emph{Promo.}\ merges two unconfirmed fragments into a newly confirmed node, recovering an otherwise lost object.}
\label{tab:critic_behavior}
\setlength{\tabcolsep}{5pt}
\small
\begin{tabular}{l c ccc c}
\toprule
\textbf{Scene} & \textbf{Calls} &\textbf{Dedup.} & \textbf{Absorb.} & \textbf{Promo.} & \textbf{Rejected} \\
\midrule
\texttt{0011\_00} & 25 & 7 & 7 & 1 & 10 \\
\texttt{0030\_00} & 34 & 3 & 4 & 0 & 27 \\
\texttt{0046\_00} & 19 & 1 & 3 & 0 & 15 \\
\texttt{0086\_00} &  9 & 0 & 2 & 1 &  6 \\
\texttt{0222\_00} & 14 & 0 & 6 & 1 &  7 \\
\texttt{0378\_00} & 14 & 2 & 6 & 0 &  6 \\
\texttt{0389\_00} & 14 & 1 & 6 & 2 &  5 \\
\texttt{0435\_00} & 15 & 3 & 4 & 0 &  8 \\
\midrule
\textbf{Overall} & \textbf{144} & \textbf{17} & \textbf{38} & \textbf{5} & \textbf{84} \\
\bottomrule
\end{tabular}
\end{table}

\begin{figure}[tb]
    \centering
    \includegraphics[width=1\linewidth]{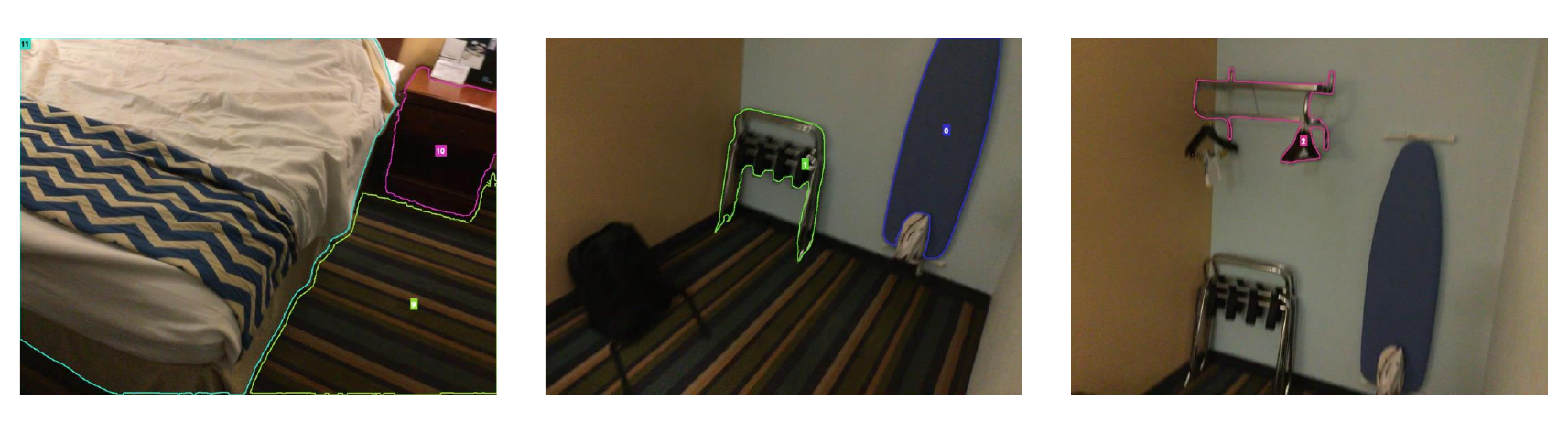}
    \caption{\textbf{Description Agent Call.} A single multi-target call on ScanNet \texttt{scene0389\_00}: three SoM-annotated frames jointly cover six objects, refining coarse labels into fine-grained descriptions in one VLM call.}
    \label{fig:desc_examples}

\end{figure}

\paragraph{Description Agent behavior.}
We illustrate the Description Agent's refinement (Sec.~3.3 of the main paper) using \texttt{scene0389\_00} from ScanNet~\cite{dai_scannet_2017}.~\cref{fig:desc_examples} shows the SoM-annotated~\cite{yang2023setofmark} frames sent to the agent along with the current track labels. The agent refines coarse labels into fine-grained descriptions. For instance, two tracks both labeled \textit{board} are disambiguated into \textit{padded ironing board (wall-mounted)} and \textit{folding luggage rack (metal with straps)}, while \textit{nightstand} becomes \textit{wooden nightstand (stained two-drawer)}. Beyond labels, the agent returns structured attributes (material, color, state, texture) that support the disambiguation of same-class instances during grounding. In one case (track\_2), the agent incorrectly overrides the correct consensus label \textit{clothes hanger} with \textit{wall-mounted reading lamp}. Since the backend consensus label and the agent's description coexist in the scene graph, incorrect descriptions do not destroy the original label.

\section{Incremental Improvement}
To demonstrate how our system improves over time, we report how the semantic segmentation scores evolve during exploration. We report metrics at four coverage checkpoints on both Replica and ScanNet (\cref{tab:coverage}). On Replica, performance improves steadily across all metrics, reaching 0.53 mAcc at just 50\% coverage, already surpassing the final scores of most batch-based baselines in Tab. 1 of the main paper. On ScanNet, early-stage metrics start lower (0.09 mIoU at 10\%) but recover quickly once diverse rooms are reached, with mIoU and f-mIoU more than doubling between 10\% and 30\% coverage.

\begin{table}[t]
  \centering
  \small
  \setlength{\tabcolsep}{3pt}
  \caption{\textbf{Incremental Convergence.} Segmentation metrics on Replica and ScanNet improve steadily as exploration progresses.}
  \label{tab:coverage}
  \begin{tabular}{@{}lccc@{\hskip 4pt}ccc@{}}
    \toprule
    & \multicolumn{3}{c}{\textbf{Replica}} & \multicolumn{3}{c}{\textbf{ScanNet}} \\
    \cmidrule(lr){2-4} \cmidrule(lr){5-7}
    \textbf{Coverage} & \textbf{mAcc} & \textbf{mIoU} & \textbf{f-mIoU} & \textbf{mAcc} & \textbf{mIoU} & \textbf{f-mIoU} \\
    \midrule
    10\% & 0.29 & 0.12 & 0.23 & 0.27 & 0.09 & 0.11 \\
    30\% & 0.45 & 0.21 & 0.42 & 0.41 & 0.21 & 0.25 \\
    50\% & 0.53 & 0.32 & 0.48 & 0.49 & 0.27 & 0.32 \\
    100\% & 0.58 & 0.37 & 0.61 & 0.75 & 0.44 & 0.46 \\
    \bottomrule
  \end{tabular}
\end{table}

\section{Qualitative Results}
\label{sec:qual_seg}
\begin{figure}[tb]
    \centering
    \includegraphics[width=1\linewidth]{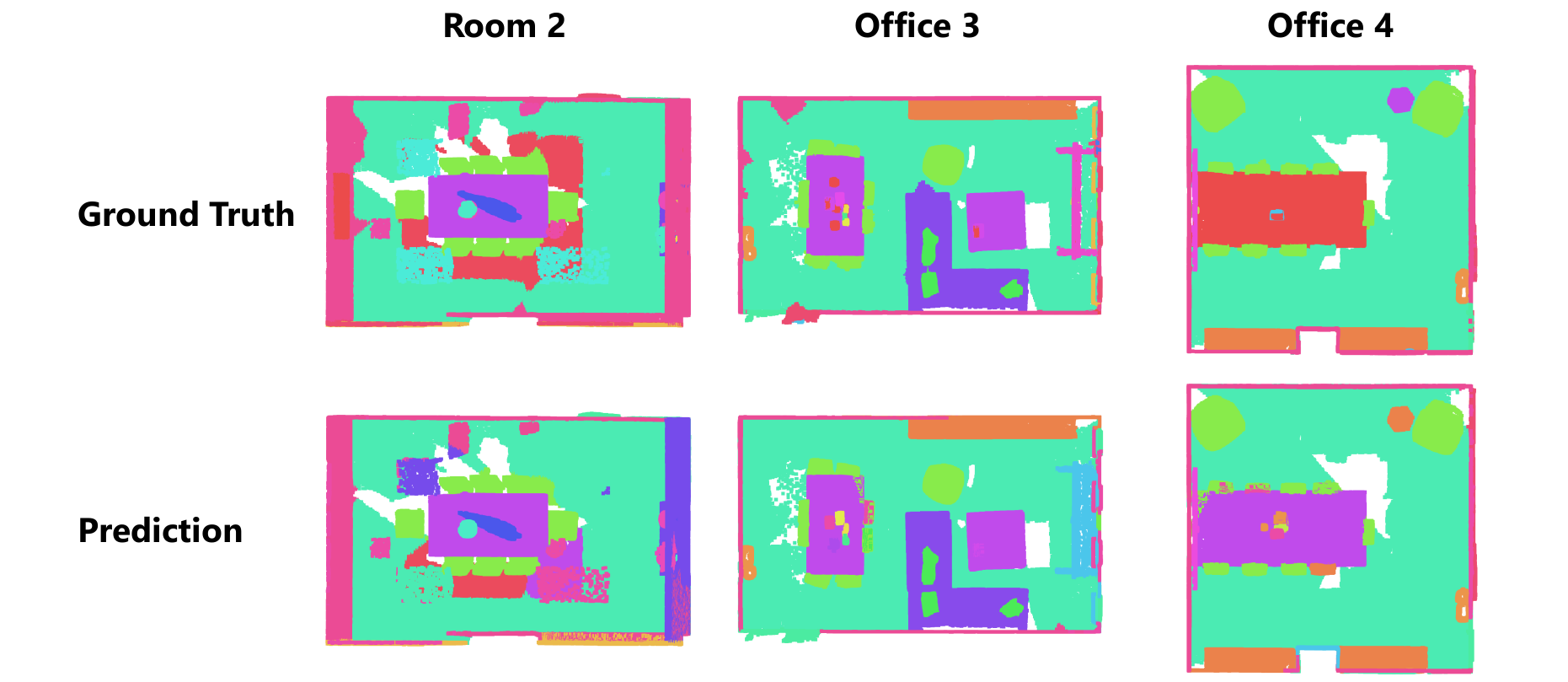}
    \caption{\textbf{Qualitative Segmentation Results.} Open-vocabulary semantic segmentation on three Replica scenes: ground truth (top), our prediction (bottom).}
    \label{fig:qual_semseg}
\end{figure}

\paragraph{Semantic segmentation.}
\cref{fig:qual_semseg} compares our open-vocabulary semantic segmentation predictions with the ground truth for three Replica scenes.
\texttt{Room 2} accurately recovers most object boundaries, although large objects such as rugs remain a known challenge for incremental association.
These objects span many frames and viewpoints, which makes it difficult to consolidate them into a single, consistent track.
\texttt{Office 3} shows strong overall coverage, but fine-grained objects (e.g. small desk items or thin wall-mounted fixtures) are occasionally missed when they fall below observation filtering thresholds or receive too few confident detections to be promoted to confirmed tracks.
\texttt{Office 4} reveals a notable evaluation artifact: objects assigned to generic ground-truth categories (\eg \emph{other}) are penalized even when our prediction is semantically plausible, such as a surface correctly identified as a table.
Across all three scenes, remaining errors concentrate on small or heavily occluded objects and large ambiguous surfaces, while the majority of each scene is segmented correctly without requiring complete reconstruction.

\begin{figure}[tb]
    \centering
    \includegraphics[width=0.99\linewidth]{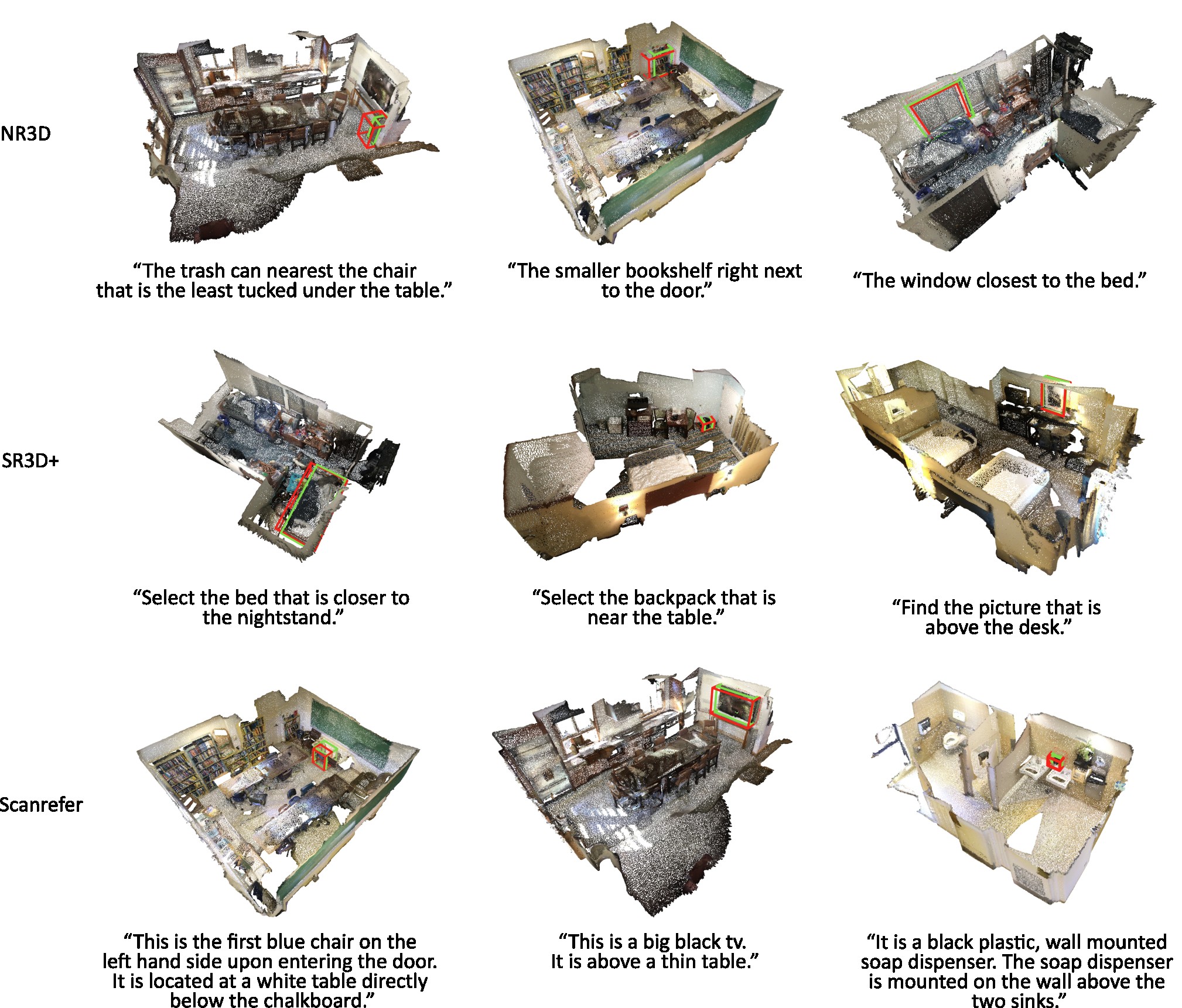}
    \caption{\textbf{Qualitative Grounding Results.} On Nr3D, Sr3D+, and ScanRefer.
    \textcolor{green!60!black}{Green} denotes the ground-truth and
    \textcolor{red!70!black}{red} the predicted bounding box. All shown examples satisfy A@0.25.}
    \label{fig:grounding_vis}
\end{figure}

\paragraph{Grounding.}
We also provide multiple examples on the grounding task in \cref{fig:grounding_vis}, covering Nr3D, Sr3D+, and ScanRefer. Each example shows the predicted bounding box alongside the ground truth for a given natural-language query.

\section{Runtime and Cost Analysis}
\label{sec:supp_runtime}

We profile end-to-end latency and VLM cost across all eight ScanNet sequences used in our temporal benchmark. All timings are reported with visualization, evaluation, and exports disabled in our latency-benchmark mode. \cref{tab:latency_scannet_all} summarizes the per-scene breakdown.

\paragraph{Online path.}
Across all scenes, the online path averages 1.45\,s per keyframe, of which the frontend contributes 1.10\,s (76\% of online time), making it the primary optimization target.

\paragraph{Asynchronous agents.}
Both VLM agents run in dedicated background threads, decoupled from per-keyframe processing.
On average, the Description Agent takes 22.7\,s to complete, with a mean lag of 28.9 keyframes (${\sim}289$ original frames at stride~10). The Critic Agent takes an average of 19.1\,s to complete, with a mean lag of 27.7 keyframes (${\sim}277$ original frames).
These lags are expected by design: the agents trade immediacy for selectivity by processing only high-value targets identified by the schedulers.

\paragraph{VLM call budget.}
Across the eight scenes, the system runs 205 VLM calls in total: 61 Description-Agent
and 144 Critic pair-verification calls (60 merges, 41.7\% acceptance). The multi-target
scheduler is what keeps enrichment cheap: a naive per-object-per-keyframe strategy would
need ${\sim}9{,}300$ calls on a single typical scene (\eg, 39~tracks $\times$
238~keyframes), whereas our Description Agent covers all targets in only 4--16 calls per
scene. The total end-to-end wall-clock time across all runs is 3{,}561\,s (59.4\,min), of
which 305\,s (8.6\%) is spent draining pending asynchronous work after the last keyframe.

\begin{table}[tb]
  \centering
  \caption{%
    \textbf{Cross-Scene Runtime.} Summary on the eight ScanNet benchmark sequences.
    Online, Front, and Back denote mean per-keyframe times in ms. D and C denote
    Description and Critic agents, reported as latency in seconds and mean lag in
    keyframes. KFs is the number of keyframes. VLM D/C/T lists the number of Description,
    Critic, and total VLM calls. Critic S/A/R reports submitted, accepted, and rejected
    Critic pairs. Wall s is the total wall-clock time in seconds. The Average row reports
    means across all eight sequences.}
  \label{tab:latency_scannet_all}
  \scriptsize
  \setlength{\tabcolsep}{4pt}
  \resizebox{\linewidth}{!}{%
  \begin{tabular}{@{}l r r r r r r r r r@{}}
    \toprule
    \textbf{Scene} & \textbf{KFs} & \textbf{Online} & \textbf{Front} & \textbf{Back} & \textbf{D s\,/\,lag} & \textbf{C s\,/\,lag} & \textbf{VLM D/C/T} & \textbf{Critic S/A/R} & \textbf{Wall s} \\
    \midrule
    \texttt{0011\_00} & 238 & 1764 & 1511 & 246 & 20.8\,/\,25.4 & 19.6\,/\,24.7 & 7\,/\,25\,/\,32 & 25\,/\,15\,/\,10 & 456.4 \\
    \texttt{0030\_00} & 250 & 1537 & 1161 & 361 & 21.6\,/\,29.7 & 18.8\,/\,27.0 & 7\,/\,34\,/\,41 & 34\,/\,7\,/\,27 & 433.1 \\
    \texttt{0046\_00} & 248 & 1441 & 1109 & 318 & 22.4\,/\,23.4 & 16.9\,/\,24.9 & 8\,/\,19\,/\,27 & 19\,/\,4\,/\,15 & 400.2 \\
    \texttt{0086\_00} & 143 & 1123 & 988 & 133 & 29.1\,/\,36.2 & 26.0\,/\,37.8 & 4\,/\,9\,/\,13 & 9\,/\,3\,/\,6 & 206.2 \\
    \texttt{0222\_00} & 541 & 1957 & 1246 & 685 & 28.7\,/\,25.3 & 20.0\,/\,21.0 & 16\,/\,14\,/\,30 & 14\,/\,7\,/\,7 & 1086.8 \\
    \texttt{0378\_00} & 190 & 1466 & 1082 & 372 & 18.3\,/\,24.4 & 18.0\,/\,25.9 & 5\,/\,14\,/\,19 & 14\,/\,8\,/\,6 & 314.7 \\
    \texttt{0389\_00} & 142 & 937 & 772 & 160 & 20.7\,/\,37.2 & 16.3\,/\,26.8 & 4\,/\,14\,/\,18 & 14\,/\,9\,/\,5 & 171.5 \\
    \texttt{0435\_00} & 328 & 1389 & 941 & 438 & 19.7\,/\,29.5 & 17.2\,/\,33.6 & 10\,/\,15\,/\,25 & 15\,/\,7\,/\,8 & 492.3 \\
    \midrule
    \textbf{Average} & \textbf{260} & \textbf{1452} & \textbf{1101} & \textbf{339} & \textbf{22.7\,/\,28.9} & \textbf{19.1\,/\,27.7} & \textbf{7.6\,/\,18.0\,/\,25.6} & \textbf{18.0\,/\,7.5\,/\,10.5} & \textbf{445.2} \\
    \bottomrule
  \end{tabular}
  }
\end{table}

\section{Frontend Design Choices}
\begin{figure}[tb]
    \centering
    \includegraphics[width=1\linewidth]{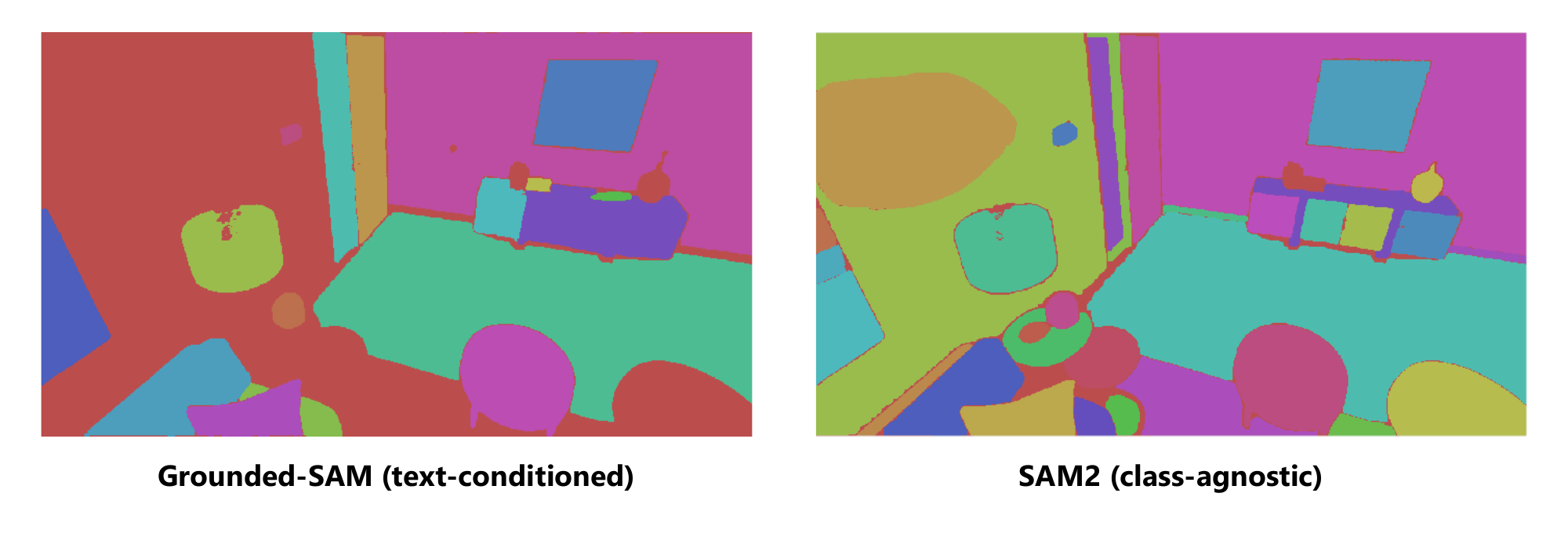}
    \caption{\textbf{Grounded-SAM vs.\ SAM2.} Grounded-SAM (left) uses text-conditioned queries and yields fewer, object-aligned masks; SAM2 (right) produces class-agnostic masks that fragment objects.}
    \label{fig:gsam_vs_sam2}
\end{figure}

\paragraph{Why text-conditioned segmentation?}
Following prior open-vocabulary scene graph methods~\cite{mei_vocabularyfree_2025,zhang2025openvocabularya}, we adopt Grounded-SAM~v2~\cite{ren2024grounded,liu_grounding_2025,ravi_sam_2024} as our segmentation frontend. This is a natural fit for our pipeline, which anchors association, label voting, and embedding selection on per-mask text labels (Eqs.~1, 3, 5 of the main paper): each mask inherits a grounded label directly from the detection stage, requiring no separate labeling step. Category-agnostic alternatives such as standalone SAM2~\cite{ravi_sam_2024} would require post-hoc labeling and tend to fragment objects into multiple regions (\cref{fig:gsam_vs_sam2}), introducing noise into downstream components.

\paragraph{Why Qwen3-VL as the proposal model?}
Grounded-SAM requires a set of textual object candidates for each frame. Since label stability has a significant downstream impact, the choice of proposal model is critical to avoid the cascade of noisy or overproduced tags throughout our backend processes: association, voting, and embedding selection.
Switching from RAM++~\cite{huang_openset_2025} to
Qwen3-VL-2B-Instruct~\cite{bai_qwen3vl_2025} yields the largest single
improvement in our backend ablation (+0.09~mAcc, Tab.~4 of the
main paper). RAM++ typically overproduces object names, introducing
irrelevant categories that destabilize incremental tracking.
Qwen produces concise, object-centric labels (\eg ``chair'', ``table'',
``shelf'') that provide a reliable anchor for the full pipeline.
Its compact size enables fully local deployment without relying on external APIs, whereas GPT-based captioning, despite being viable, introduces substantial latency that is unsuitable for per-frame operation.

{\raggedbottom
\section{Prompts}
\label{sec:prompts}
We provide the exact prompts used for the frontend tagger, asynchronous VLM agents, and grounding evaluator.

\begin{tcolorbox}[
  colback=gray!5, colframe=gray!50,
  title=Frontend Tagging Prompt,
  beforeafter skip=20pt
]
\scriptsize\ttfamily
\textbf{--- User ---}\\[2pt]
List the distinct objects you can see in this scene. Respond with a comma-separated list of short nouns only.
\end{tcolorbox}

\begin{tcolorbox}[
  colback=gray!5, colframe=gray!50,
  title=Critic Agent Prompt,
  beforeafter skip=20pt
]
\scriptsize\ttfamily
\textbf{--- System ---}\\[2pt]
You are an expert 3D Scene Graph Critic.
Your task is to resolve redundant tracks in a 3D scan.
You will receive one or two images.
If there is one image, both objects are shown together in that same image.
If there are two images, Image 1 is Object A and Image 2 is Object B.
Decide if they should be MERGED into one entity or KEPT as distinct objects.
Return JSON only.\\[6pt]
\textbf{--- User ---}\\[2pt]
Two candidate tracks for merge review:\\
- Object A: ID=\{id\_a\}, label='\{label\_a\}', observations=\{match\_count\_a\}\\
- Object B: ID=\{id\_b\}, label='\{label\_b\}', observations=\{match\_count\_b\}\\
~~(confirmation threshold: \{confirm\_threshold\}; tracks below it will likely be pruned)\\
Image count: \{pair\_image\_count\}\\
Semantic cosine (SBERT): \{pair\_semantic\_cosine\}\\[4pt]
Geometry signals:\\
- Containment: \{pair\_containment\_summary\}\\
- Offset (m, Z=vertical): \{pair\_offset\_summary\}\\
- Horizontal / Vertical offset: \{pair\_horizontal\_offset\_m\} / \{pair\_vertical\_offset\_m\}\\
- B/A volume ratio: \{pair\_relative\_size\_ratio\_b\_to\_a\} (\{pair\_relative\_size\_summary\})\\
- Viewpoint mismatch cost: \{pair\_pov\_cost\}\\
- Frames: A=\{frame\_a\}, B=\{frame\_b\}\\[4pt]
MERGE when:\\
- Same physical object from different views (duplicate)\\
- One is a sub-part/fragment of the other\\
- Single object split by occlusion\\[4pt]
KEEP when:\\
- Distinct instances, even if same class and adjacent\\
- Different color/texture/material\\
- Objects only touch at an edge\\
- Small object merely inside larger bbox (containment != same instance)\\
- Either label is a wall variant\\
- Evidence is ambiguous -> default KEEP\\[4pt]
Evaluate in order: appearance -> geometry -> containment -> wall rule -> uncertainty.\\
Output JSON only (keys in this order):\\
\{\\
~~"reasoning": "<concise explanation>",\\
~~"action": "MERGE" or "KEEP"\\
\}
\end{tcolorbox}

\begin{tcolorbox}[
  colback=gray!5, colframe=gray!50,
  title=Description Agent Prompt,
  breakable,
  beforeafter skip=20pt
]
\scriptsize\ttfamily
\textbf{--- System ---}\\[2pt]
You are a precise computer vision assistant specializing in fine-grained
object attribute extraction. Your goal is to generate a structured JSON
description of objects grounded strictly in their visual appearance.\\[6pt]
\textbf{--- User ---}\\[2pt]
The attached images show a scene with specific objects marked by numeric
ID tags (e.g., [14]). Focus ONLY on the objects listed below. Aggregate
visual evidence from all provided views to ensure completeness.\\[4pt]
Target Objects: \{ids\}\\[4pt]
Task:\\
For each target ID, return a JSON object with:\\
1. "label": A specific, fine-grained noun (e.g., 'winged armchair' instead
   of 'chair', 'ceramic vase' instead of 'decor').\\
2. "attributes":\\
~~~- "material": (e.g., wood, velvet, glass, plastic)\\
~~~- "color": (Dominant colors)\\
~~~- "state": (e.g., open, closed, folded, dirty, wet, empty)\\
~~~- "texture": (e.g., glossy, matte, rough, knitted)\\[4pt]
Constraint:\\
- Do NOT hallucinate attributes not visible in the images.\\
- If an object is completely unclear, set attributes to null.\\
- Output pure JSON mapping \{ "id": \{ "label": "...", "attributes": \{...\} \} \}.\\
- The numeric ID tags are only identifiers; ignore the digits/label text and
  use only the pixels inside the ID contour/overlay.\\[6pt]
\textbf{--- Prior append (when available) ---}\\[2pt]
Tracker Memory (weak prior):\\
We keep a running scene graph estimate for each object ID. Use this as slight
guidance only. Priors may be stale or inaccurate. If visual evidence disagrees,
override the prior label/attributes/description.\\
Prior snapshot per target ID:\\
\{prior\_lines\}\\[2pt]
\textrm{\textit{Each prior line is formatted as:}}\\
- [3] label="armchair"; description="brown chair near wall";
  attributes=\{"color":["brown"],"material":["fabric"]\}
\end{tcolorbox}


\begin{tcolorbox}[colback=gray!5, colframe=gray!50,
  title=Grounding Evaluator Prompt,
  breakable, beforeafter skip=10pt]
\scriptsize\ttfamily
\textbf{--- System ---}\\[2pt]
You are a helpful assistant. The user provides a 3D scene represented by:\\
-- a list of objects, each containing an identifier, descriptive fields, and optional
spatial, semantic, or other attributes\\
-- a list of edges, each describing relationships between objects, with identifiers,
relation labels, and optional metrics\\
Your task is to identify the object best matching the user's query, based on this
structured information. Follow these answering guidelines strictly:\\
1. Select the object ID that best matches the query.\\
2. Provide a short, factual explanation.\\[6pt]
\textbf{--- User ---}\\[2pt]
Below is a 3D scene composed of objects and edges. Each edge is represented as
[subject\_id, predicate, object\_id, distance\_m (optional), compass (optional)].\\
\{scene\_line\}\\
\{edge\_line\}\\
query=\{utterance\}
\end{tcolorbox}

}

\section{Full implementation details}

For completeness,~\cref{tab:hyperparams_ext} summarizes the key implementation parameters used in our system beyond those reported in the main paper. The released code will provide the complete configuration and default settings.

\begin{table}[tb]
\centering
\caption{\textbf{Implementation Parameters.} Extended hyperparameters beyond those in the main paper.}
\label{tab:hyperparams_ext}
\small
\resizebox{\linewidth}{!}{%
    
    \begin{tabular}{@{}l@{\hspace{0.7cm}}l@{\hspace{0.7cm}}l@{}}
    \toprule
    \textbf{Component} & \textbf{Parameter} & \textbf{Value} \\
    \midrule
    Frontend  & Label proposer & Qwen/Qwen3-VL-2B-Instruct \\
            & Segmentation model & Grounded-SAM~2 \\
            & Grounding-DINO variant & Swin-T (OGC) \\
            & SAM2 variant & sam2.1\_hiera\_base\_plus \\
            & Qwen3-VL max tokens & 128 \\
            & Grounding-DINO box / text threshold & 0.35 / 0.25 \\
            & SAM2 mask threshold / max proposals & 0.5 / 100 \\
            & Prompt budget (max labels / Qwen tags) & 64 / 40 \\
    \midrule
    Observation Filtering & Erosion kernel / min area & $5\times5$ / 100 px \\
    & DBSCAN eps / min\_samples / voxel & 0.05\,m / 7 / 0.025\,m \\
    & Tracker min segment area / confidence & 30 px / 0.1 \\
    & Min points after filter / cluster ratio & 30 / 0.7 \\
    \midrule
    Backend & CLIP backbone & EVA02-base \\
    & CLIP crop padding & 30 px \\
    & Frame stride (Replica / ScanNet runs) & 5 / 10 \\
    & Semantic eval prompt template & \texttt{an image of \{label\}} \\
    \midrule
    Spatial Edges & Rule-based proximity threshold & 2.0\,m \\
    & Top-k nearest edges & 5 \\
    \midrule
    VLM Agents (shared) & Model & \texttt{gpt-5-mini} \\
    & Max output tokens & 5000 \\
    & Critic reasoning effort & \texttt{high} \\
    \midrule
    Description Agent & Target selection (obs / area / growth / interval) & 5 / 2500 px / 0.3 / 2 s \\
    & Description agent batch / concurrency / queue & 5 / 4 / 32 \\
    & Multi-target scheduler window / batch budget & 30 / 3 \\
    & Multi-target scheduler gates & cov. 0.8; occ. 0.02; res. gain 0.2 \\
    & Utility scale / background weight & 100 / 0.25 \\
    \midrule
    Critic Agent & Critic concurrency / queue & 30 / 64 \\
    & SBERT model & \texttt{all-MiniLM-L6-v2} \\
    & Pair candidate scope / trigger & \texttt{all} / 2 matches \\
    & Pair semantic / IoU gate & 0.8 / 0.01 \\
    & Pair view budget / overlay mode & 30 / \texttt{mask\_pair} \\
    & False-negative queue max pairs / flush & 20 / 50 frames \\
    \bottomrule
    \end{tabular}
    }
\end{table}

\end{document}